\documentclass[acmsmall]{acmart}

\usepackage{color}

\usepackage{booktabs}
\usepackage{amsmath}

\usepackage{amssymb}
\usepackage{multirow}
\usepackage{cases}

\usepackage{algorithm}
\usepackage{algorithmicx}
\usepackage{algpseudocode}

\AtBeginDocument{%
  \providecommand\BibTeX{{%
    \normalfont B\kern-0.5em{\scshape i\kern-0.25em b}\kern-0.8em\TeX}}}

\setcopyright{acmcopyright}
\copyrightyear{2021}
\acmYear{2021}
\acmDOI{00.0000/0000000.0000000}

\acmJournal{TOMM}
\acmVolume{0}
\acmNumber{0}
\acmArticle{0}
\acmMonth{0}

\newcommand{\bfstart}[1]{\noindent\textbf{#1.}}
\newcommand{\ie}{\textit{i}.\textit{e}.}

\begin{document}

\title{Answer Questions with Right Image Regions: A Visual Attention Regularization Approach}


%\authornotemarks[1]

\author{Yibing Liu}
\email{lyibing112@gmail.com}
\author{Yangyang Guo}
\email{guoyang.eric@gmail.com}
\author{Jianhua Yin}
\email{jhyin@sdu.edu.cn}
\author{Xuemeng Song}
\email{sxmustc@gmail.com}
\affiliation{%
	\institution{Shandong University}
	\department{School of computer science and technology}
	\streetaddress{72 Binhai Road, Jimo}
	\city{Qingdao}
	\state{Shandong Province}
	\country{China}
	\postcode{266237}
}
\author{Weifeng Liu}
\email{liuwf@upc.edu.cn}
\affiliation{%
	\institution{China University of Petroleum (East China)}
	\streetaddress{66 West Changjiang Road, Huangdao District}
	\city{Qingdao}
	\state{Shandong Province}
	\country{China}
	\postcode{266580}
}
\author{Liqiang Nie}
\email{nieliqiang@gmail.com}
\affiliation{%
	\institution{Shandong University}
	\streetaddress{72 Binhai Road, Jimo}
	\city{Qingdao}
	\state{Shandong Province}
	\country{China}
	\postcode{266237}
}

\authorsaddresses{%
Y. Liu, Y. Guo, J. Yin, X. Song, and L. Nie are with the School of Computer Science and Technology, Shandong University, 72 Binhai Road, Jimo, Qingdao, Shandong Province, China (266237); emails: lyibing112@gmail.com, guoyang.eric@gmail.com, jhyin@sdu.edu.cn, sxmustc@gmail.com, nieliqiang@gmail.com; L. Nie is the corresponding author (emails: nieliqiang@gmail.com); W. Liu is with the College of Control Science and Engineering, China University of Petroleum (East China), 66 West Changjiang Road, Huangdao District, Qingdao, Shandong Province, China (266580); email: liuwf@upc.edu.cn.
}

%No.1500, Shunhua Road, Jinan, Shandong Province, P.R. China, 250101; emails:

\renewcommand{\shortauthors}{Y. Liu et al.}

\begin{abstract}
	{Visual attention in Visual Question Answering (VQA) targets at locating the right image regions regarding the answer prediction, offering a powerful technique to promote multi-modal understanding. }
	However, recent studies have pointed out that the highlighted image regions from the visual attention are often irrelevant to the given question and answer, leading to model confusion for correct visual reasoning. 
	To tackle this problem, existing methods mostly resort to aligning the visual attention weights with human attentions. 
	Nevertheless, gathering such human data is laborious and expensive, making it burdensome to adapt well-developed models across datasets. 
	To address this issue, in this paper, we devise a novel visual attention regularization approach, namely AttReg, for better visual grounding in VQA. 
	Specifically, AttReg firstly identifies the image regions which are essential for question answering yet unexpectedly ignored (\textit{i.e.}, assigned with low attention weights) by the backbone model. And then a mask-guided learning scheme is leveraged to regularize the visual attention to focus more on these ignored key regions. 
	The proposed method is very flexible and model-agnostic, which can be integrated into most visual attention-based VQA models and require no human attention supervision. 	Extensive experiments over three benchmark datasets, \textit{i.e.}, VQA-CP v2, VQA-CP v1, and VQA v2, have been conducted to evaluate the effectiveness of AttReg. 
	As a by-product, when incorporating AttReg into the strong baseline LMH, our approach can achieve a new state-of-the-art accuracy of {60.00\%} with an absolute performance gain of {7.01\%}  on the VQA-CP v2 benchmark dataset.
	In addition to the effectiveness validation, we recognize that the faithfulness of the visual attention in VQA has not been well explored in literature. In the light of this, we propose to empirically validate such property of visual attention and compare it with the prevalent gradient-based approaches. 
\end{abstract}





\maketitle
\newpage
\section{Introduction}
{
With the great progress of natural language processing, computer vision, and multimodal representation learning, Visual Question Answering (VQA) has emerged as a significant interdisciplinary task in recent years. VQA aims to correctly answer natural language questions about an image \cite{Dataset:VQA_v1,AttentionMethods:BTDP,AttentionMethods:Multilevel_VQA} or a video \cite{MM:Video1,MM:Video2,MM:Video3}. As an ``AI-complete'' problem, VQA encounters a variety of research challenges, such as recognition, counting, and multi-modal understanding.
}
Canonical methods often cast VQA as a classification task \cite{Baseline:NMN,loss_rescaling,AttentionMethods:HieCoAtt,AttentionMethods:MCB,AttentionMethods:SAN,AttentionMethods:UpDn,AttentionMethods:HardAtt,AttentionMethods:Sem_VQA,Baseline_LP:Quantify_LP,adaVQA}, where the image and question are processed via Convolutional Neural Networks (CNNs) and Recurrent Neural Networks (RNNs), respectively. Among the existing methods, an intriguing design is to apply the visual attention mechanism to image regions based on the given question, equipping VQA models with the capability of visual grounding and explanation.

\begin{figure*}
	[t]
	\centering
	\includegraphics[width=0.7\columnwidth]{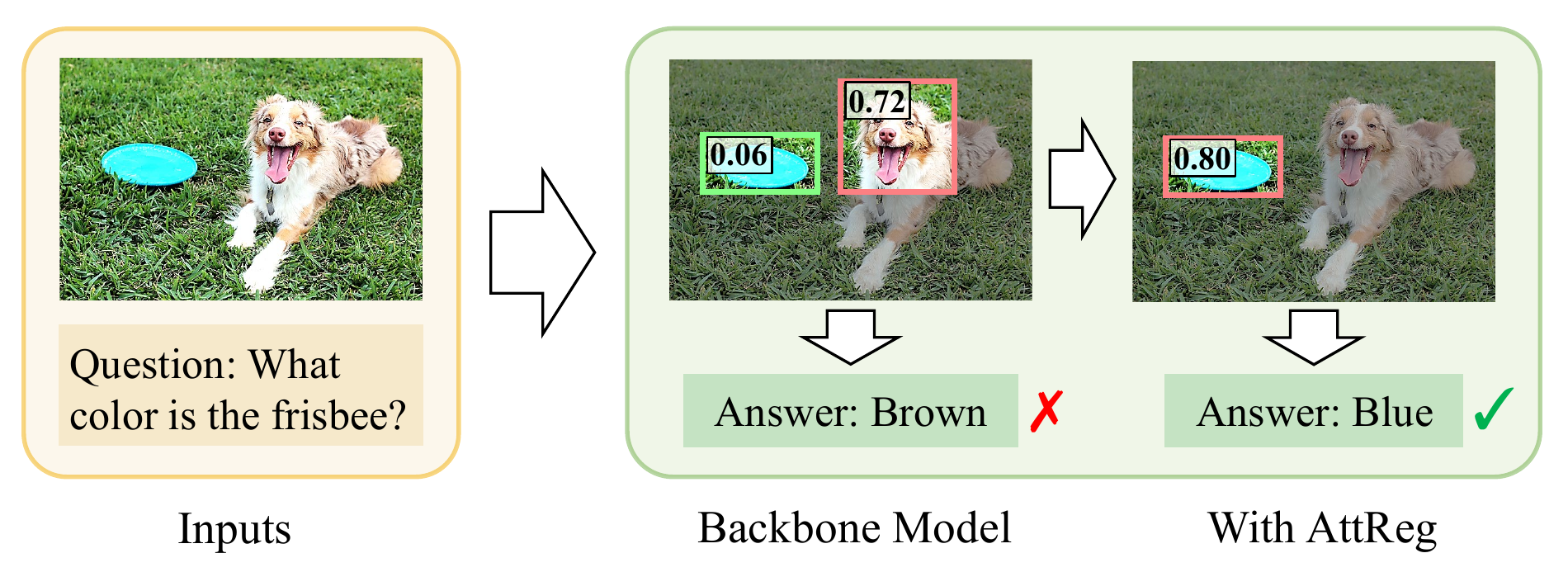} 
	\caption{ 
		Demonstration of the visual attention in VQA and our regularization method. The green box (\textit{i.e.}, \textit{frisbee}) in the middle image denotes the key region ignored by the backbone visual attention mechanism, wherein lower attention weight is assigned compared to the pink box (\textit{i.e.}, \textit{dog}). In the rightmost image, our AttReg regularizes the {model} to focus more on the most relevant \textit{frisbee} region. 
	}
	\label{intro}
\end{figure*}

Generally speaking, the visual attention in VQA executes an explicit step of identifying ``where to look'' \cite{AttentionMethods:Wheretolook,VQA_Survey}. 
To be more specific, it allows the model to assign distinctive weights to different image regions, which are computed via the semantic similarity between the given question and image features, as illustrated in Figure \ref{intro}. 
Similar to human vision systems, the image regions with high attention weights are commonly deemed as where the model looks at when making {predictions}.
Accordingly, by spotting such relevant image regions, the visual attention is able to not only reduce noisy features but also construct more refined visual representations.

%However, even today's VQA models hugely benefit from the visual attention mechanisms, the ``faithfulness'' of generated attention maps has not been well explored in VQA to our knowledge. 

%Although many researchers have realized this issue and developed a set of approaches
%To deal with this issue, a seemingly straightforward solution is to provide explicit human attention supervision for attention weights learning -- aligning image region importance with the human attention (Qiao, Dong, and Xu 2018; Zhang, Niebles, Soto 2019). 
%Although this issue is commonly , this problem is non-trivial to deal with. One reason for this intractability is the irresistible language biases (\textit{i.e.,} ), which 
%For instance, since the answer \textit{2} is the most frequent answer in the question begin with \textit{how many}, VQA models may tend to exploit this easy-to-learn shortcut to predict the answer and ignore the appropriate visual guidance. 
Despite of the fact that existing VQA models have benefited a lot from the visual attention, one imperative issue lies in the lack of guidance for visual grounding. 
This often leads visual attention mechanisms to focus on image regions which are less relevant to the correct answer \cite{Baseline_AttSupv(Argument):HINT,Dataset:HAT}. 
As shown in Figure \ref{intro}, the visual attention in the backbone model focuses on the much less important region \textit{dog} while ignores the most relevant one \textit{frisbee},  misleading the model to predict the incorrect answer \textit{brown}. 
To deal with this issue, a seemingly straightforward solution is to align visual attention weights with explicit human attentions \cite{AttentionSupervision:HAN,AttentionSupervision:InterpretableVQA}. 
Collecting such data is, however, expensive and difficult \cite{Dataset:HAT}, thereby limiting the practicality of this kind of approach. 

{
Orthogonal to the visual attention, another prevalent branch for improving visual grounding is resorting to gradient-based techniques (specifically Grad-CAM \cite{Tech:GradCAM}), which allow us to understand model prediction through activation mappings from gradients.
}
For instance, the method in \cite{Baseline_AttSupv(Argument):HINT} calculates the gradient of each image region according to { the predicted scores of ground-truth answers}, and encourages the rank consistency between the gradients and human attentions for better visual grounding. More recently, Wu and Mooney \cite{Baseline_AttSupv:SCR} utilized the textual annotations (\textit{e.g.}, QA pairs) as auxiliary information and {regularized} visual grounding by penalizing the gradients of important regions to wrong answers.
However, our experiments reveal that {such explanations obtained via Grad-CAM} are not trustworthy for visual grounding compared with that of visual attention. 
{One notable problem is that the image regions with large gradients are not positively related to model predictions}, which violates the intuition that Grad-CAM is specially designed for visual grounding in VQA~\cite{Baseline_AttSupv(Argument):HINT}.

%Another plausible way is to employ Gradient-weighted Class Activation Mapping (Grad-CAM) \cite{Tech:GradCAM} as the visual explanations to guide the visual grounding. They derive the local contribution of each image region to correct answers and encourage the rank consistency between the obtained region gradient and human annotations \cite{Baseline_AttSupv:SCR,Baseline_AttSupv(Argument):HINT,Baseline_AttSupv:CSS}.

%Specifically, for each training sample, we firstly identify the ignored key regions, \textit{i.e.}, the image regions highly related to the QA pair but assigned lower attention weights. 
%Specifically, for each training sample, we firstly identify the ignored key regions, \textit{i.e.}, the image regions essential for the question answering but assigned lower attention weights. 

To overcome the limitations of previous approaches, in this paper, we present a novel visual attention regularization method, dubbed as AttReg, to guide the learning of visual attention in VQA. 
As illustrated in Figure \ref{intro}, AttReg aims to achieve better accuracy via attention weights regularization. 
{Concretely, for each training sample, AttReg first identifies the ignored key regions, \textit{i.e.}, the image regions pivotal to the question answering yet assigned with low attention weights.
Then, AttReg constructs a curated image with these ignored key regions being masked, and a new training sample is thereby built by pairing the same question to the curated image.
Next and crucially, different from the original training, AttReg forces the model to answer \textit{None} when given the curated sample since the relevant image regions (\textit{i.e.}, ignored key regions) are masked and invisible under this situation\footnote{{\emph{None} can be understood as ``I don't know''. That is, a model's predicted scores of all candidate answers are zeros.}}.
In this way, the model would be regularized to shift more attention weights to these ignored key regions, hence boosting the answer prediction accuracy.
}
%Then, AttReg builds another training sample via pairing the same question to a new image (curated image) whose ignored key regions are masked.
%By refraining the curated samples from being correctly answered, these ignored key regions are guided to yield stronger influence towards model predictions, thereby the model is regularized to attach higher attention weights to them and boosting the answer prediction accuracy.
%Specifically, for each training sample, AttReg firstly identifies the ignored key regions, \textit{i.e.}, the image regions pivotal to the question answering but assigned with low attention weights. 
%Thereafter, AttReg builds another sample composed of the same question and a new curated image with these ignored key regions being masked. Crucially, the model is trained to answer \textit{None} when given the curated sample since the relevant image regions are masked and invisible under this situation. 
%By refraining the curated samples from being correctly answered, these ignored key regions are guided to yield stronger influence towards model predictions, thereby the model is regularized to attach higher attention weights to them and boosting the answer prediction accuracy.
%{\color{orange}Then the advanced prediction accuracy can be achieved with such improvement in visual attention.}
In a nutshell, our end-to-end approach is simple to implement, which requires no human supervision and can be applied to most of the existing visual attention-based VQA models.
We conduct extensive experiments on three VQA datasets, \textit{i.e.}, VQA-CP v1 \cite{Dataset:VQA_CP(GVQA)}, VQA-CP v2 \cite{Dataset:VQA_CP(GVQA)}, and VQA v2 \cite{Dataset:VQA_v2}, to verify the effectiveness of the proposed method. Experimental results have shown that our proposed method is capable of guiding the visual attention learning, and enhancing the performance of many visual attention-based VQA models.

%of the visual attention (i.e., accuracy improvement)
In addition to the effectiveness validation, {we notice} that the faithfulness of the visual attention in VQA has not been well explored so far.
The faithfulness here refers to the consistency between the attention weight and the contribution of its corresponding image region to model {predictions}. 
In particular, weak faithfulness can be obtained if the regions with high attention weights often have little influence on model predictions.
To justify such a property, we empirically conduct occlusion studies \cite{Tech:GradCAM} on a well-devised visual attention model -- UpDn \cite{AttentionMethods:UpDn}. {We found that the learned attention weight is reasonably correlated} with the influence of image regions {in model prediction}, which demonstrates the favorable faithfulness of visual attention.
%By ``faithfulness'', we mean the consistency between the magnitude of attention weight and the importance of the corresponding image region towards model decisions.

In summary, the contributions of this work are three-fold:
\begin{itemize}
	\item We present a novel visual attention regularization approach in VQA, which is able to guide the model towards correctly answering questions based on right image regions. The proposed regularization method is model-agnostic, requiring no human attention supervision, and can be incorporated into most visual attention-based VQA models, such as UpDn and LMH \cite{Baseline_LP:LMH}.
	
	\item We empirically study the faithfulness of prevalent visual attention in VQA. The results exhibit that the visual attention is more faithful to model {predictions} in comparison with {Grad-CAM}.
	
	\item Extensive experiments demonstrate that the proposed method can simultaneously improve the visual grounding accuracy and the backbone model performance. As a side product, by introducing our regularization method to a strong model LMH, we can achieve a new state-of-the-art performance over the VQA-CP v2 dataset. We have released the involved data, codes, and parameter settings to facilitate other researchers in this community\footnote{\url{https://github.com/BierOne/VQA-AttReg}.}.
	
\end{itemize}

% We empirically study the faithfulness of prevalent {visual grounding tools (\textit{i.e.}, visual attention and Grad-CAM)} in VQA.

% In Section \ref{Sec_VAttExp}, we study the faithfulness of the visual attention mechanism and compare it with the Grad-CAM methods.

{The remainder of this paper is organized as follows. Section \ref{Sec_RelatedWork} reviews the related work. In Section \ref{Sec_Method}, we introduce the visual attention in VQA and our proposed method. Experimental settings and results are illustrated in Section \ref{Sec_exp}, followed by conclusion and future works in Section \ref{Sec_Conclusion}.}

\section{Related Work}
\label{Sec_RelatedWork}
In this section, we discuss two categories which are closely related to this work: visual attention in VQA and visual grounding enhancement for VQA.

\subsection{Visual Attention in VQA}
{
Towards multimodal understanding, visual attention has been widely applied in many applications, such as image captioning \cite{MM:Caption2,MM:Caption}, visual dialogue \cite{MM:dialogue}, and video action recognition \cite{MM:action}. In VQA, the visual attention is firstly introduced to }address the ``where to look'' problem by directly calculating the semantic similarity between the question and image regions \cite{AttentionMethods:Wheretolook}. 
To obtain more fine-grained visual representation, \cite{AttentionMethods:HieCoAtt} later applied the hierarchical question structure modeling for image attention. 
The question representations are learned from multi-levels, \textit{e.g.}, word-level and phrase-level, which are effectively employed to conduct visual attention recursively.
In addition, the multi-glimpse visual attention mechanism \cite{AttentionMethods:SAN} is subsequently proposed to iteratively infer the answer by attending on visual features multiple times.
Different from the above top-down visual attention methods, Anderson et al. \cite{AttentionMethods:UpDn} presented a combined bottom-up and top-down attention mechanism (UpDn), which firstly detects common objects and attributes in an image and then leverages visual attention to attend on these high-level concepts. This technique has been widely adopted and extended in recent studies \cite{Baseline_LP:RuBi, Baseline_AttSupv:SCR, Baseline_AttSupv:CSS}, boosting the performance of a series of VQA models. 

However, the above visual attention-based models are all restricted by the lack of guidance for visual grounding \cite{Dataset:HAT,Baseline_AttSupv(Argument):HINT}. 
{It is thus normal that these VQA models} are inclined to focus on irrelevant visual contents or even resort to superficial biases to answer questions~\cite{Baseline_LP:AdvReg,Baseline_AttSupv(Argument):HINT,Baseline_AttSupv:CSS}. To address this problem, in this work, we present a novel regularization strategy to guide the attention learning and boost the answer prediction accuracy.

\subsection{Visual Grounding Enhancement for VQA}
One of the most desired properties for VQA systems is to equip correct visual grounding for model {predictions}, \textit{i.e.}, right for right reasons \cite{RightForRightReason, Baseline_AttSupv:SCR}.
%Existing approaches for this purpose can be roughly classified into two groups: 1) visual attention; and 2) Grad-CAM ones.
{Existing approaches for this purpose are mainly based on two powerful techniques: 1) visual attention; and 2) Grad-CAM.
The methods based on visual attention} \cite{AttentionSupervision:HAN, AttentionSupervision:InterpretableVQA} target at aligning the attention weights with explicit human attentions. Nevertheless, since the human attention data are far too expensive and difficult to collect, the practicability and generalization of such methods are largely limited.

%Orthogonal to the visual attention, Gradient-weighted Class Activation Mapping (Grad-CAM) methods leverage the gradient value of each image region according to the model predictions to achieve visual grounding. 
%In contrast, methods in the second group utilize Grad-CAM to encourage the consistency between the gradient magnitude of each image region for the ground-truth answer and human annotations. 

{In contrast, Grad-CAM ones encourage the consistency between human annotations and region gradients derived from model predictions.}
For example, Selvaraju et al. \cite{Baseline_AttSupv(Argument):HINT} presented a human importance-aware network tuning (HINT) to enforce {the region's gradient to share the same ranking with human attentions}. 
SCR \cite{Baseline_AttSupv:SCR} leverages the textual annotations (\textit{e.g.}, QA pairs) to relate with the influential regions in images first, and then {criticizes the sensitivity of incorrect answers to these influential regions.}
More recently, Chen et al. \cite{Baseline_AttSupv:CSS} designed a Counterfactual Samples Synthesizing (CSS) scheme, {which guides visual grounding by forcing important regions to obtain high gradients from correct answers.}
Different from the above ones, Patro et al. \cite{Baseline_AttSupv:ExpVsAtt} devised an adversarial learning strategy, {which utilizes the Grad-CAM results as the surrogate supervision for attention maps.}
However, our experiments demonstrate that {the regions highlighted by Grad-CAM are not always where the model focuses on when making predictions}, which limits its visual grounding capability to some extent.
{In addition, \citet{R1:negative_analysis} pointed out that the improvement brought by these methods does not actually emerge from proper visual grounding, but from regularization effects. \citet{R2:OOD} further extended this finding with detailed analysis on in-domain and out-of-distribution (OOD) sets.}

\section{Approach}
\label{Sec_Method}
Our approach aims at regularizing VQA models to predict the right answers based on the right image regions. In the following, we first state the basic knowledge of VQA and its visual attention variant in {Section \ref{approach_preliminaries}}. The details of the proposed method are then elaborated in {Section \ref{approach_AttReg}}.

\subsection{Preliminaries}
\label{approach_preliminaries}
%$\boldsymbol{Q}$ 
\bfstart{Problem Formulation}
%\bfstart{Problem Formulation}
The goal of VQA is to provide a correct answer $\hat{A}$ to a given textual question $Q$ upon an image $I$. And the common function of VQA is formulated as a classification problem:

\begin{equation}
{\hat{A}} = \mathop{\arg\max}\limits_{A_j\in\mathcal{A}} P(A_j|Q, I; \Theta),
\end{equation}
where $\mathcal{A}$ denotes the candidate answer set and $\Theta$ denotes all the model parameters. 
{Note that in an open-ended VQA task, there can be multiple correct answers for each instance. Hence, we formulate the VQA task as a multi-label classification problem in this paper.}

%The goal of VQA is to provide a correct answer $\hat{A}$ to a given textual question $Q$ upon an image $I$. And the common function of VQA is formulated as a classification problem:
%\begin{equation}
%{\hat{A}} = \mathop{\arg\max}\limits_{A_j\in\mathcal{A}} P(A_j|Q, I; \Theta),
%\end{equation}
%where $\mathcal{A}$ denotes the candidate answer set and $\Theta$ denotes all the model parameters.

\bfstart{Visual Attention in VQA (VAtt)}
%$\boldsymbol{V} = \{v_1, ..., v_{|\boldsymbol{V}|}\}$		
%$\boldsymbol{V}\in \mathbb{R}^{d\times K}$
%$\boldsymbol{V} = \{v_k\}_{k=1}^K$
%$v_k$
{Traditional visual attention mechanisms \cite{ref1:visual7w,ref2:focalVAtt,AttentionMethods:SAN,AttentionMethods:HieCoAtt} in VQA often perform on the equal-sized image regions.}
The complementary bottom-up attention mechanism is therefore proposed to detect objects and attributes in images for identifying high-level concepts \cite{AttentionMethods:UpDn}. In this {paper}, we mainly recap the bottom-up and top-down (UpDn) attention model.
%We take the UpDn model as an example to recap how visual attention performs in VQA due to its excellent performance.

%For each image $I$, the UpDn utilizes the object detection techniques (\textit{e.g.}, Faster R-CNN \cite{Tech:RCNN}) to extract $K$ object features: $\{\boldsymbol{v}_k\}_{k=1}^K$. 
As illustrated in Algorithm \ref{alg:UpDn}, for each image $I$, the UpDn model utilizes the object detection techniques (\textit{e.g.}, Faster R-CNN \cite{Tech:RCNN,MM:RCNN}) to extract a object feature set $\mathcal{V} = \{\boldsymbol{v}_k\}_{k=1}^K$. 
And for each question $Q$, {an} RNN (\textit{e.g.}, GRU \cite{Tech:GRU}) is used to capture the sequential features represented by $\boldsymbol{q}$, \textit{i.e.} $\boldsymbol{q} = {\rm RNN}(Q)$. 
The visual attention mechanism is then utilized to refine visual representations by employing the question feature to attend on each object in the image,

{
\begin{equation}
{s}_i =  \boldsymbol{w}_a^T f_a([\boldsymbol{v}_i, \boldsymbol{q}]),
\end{equation}
\begin{equation}
\boldsymbol{\alpha} =  {\rm softmax}( \boldsymbol{s} ),
\end{equation}
\begin{equation}
\hat{\boldsymbol{v}} = \sum_{i=1}^K {\alpha}_i \boldsymbol{v}_i,
\end{equation}
}

%\begin{subequations}
%\begin{numcases}{}
%{s}_i =  \boldsymbol{w}_a^T f_a([\boldsymbol{v}_i, \boldsymbol{q}]),\\
%\boldsymbol{\alpha} =  {\rm softmax}( \boldsymbol{s} ),\\
%\hat{\boldsymbol{v}} = \sum_{i=1}^K {\alpha}_i \boldsymbol{v}_i,
%\end{numcases}
%\end{subequations}

\noindent { where} $\boldsymbol{\alpha}$ and $\boldsymbol{s}$ respectively denote the attention weights and the computed scores for image objects, $\boldsymbol{w_a}$ is a trainable parameter vector, $f_a(\cdot)$ denotes the fusion function, and $\hat{\boldsymbol{v}}$ is the final visual representation.
%The weighted visual feature $\hat{\boldsymbol{v}}$ and question feature $\boldsymbol{q}$ are then fed into the answer prediction module $f_p$ to predict the confidence for each candidate answer $A_j$ in $\mathcal{A}$,
The weighted visual feature $\hat{\boldsymbol{v}}$ and the question feature $\boldsymbol{q}$ are then fed into the answer prediction module $f_p$ to predict the confidence for all candidate answers,
\begin{equation}
\label{org_pred}
\boldsymbol{p} = f_p(\hat{\boldsymbol{v}}, \boldsymbol{q}).
\end{equation}

%\begin{equation}
%\label{org_pred}
%{p_j} = f_p(\hat{\boldsymbol{v}}, \boldsymbol{q}).
%\end{equation}

%In fact, each question may have several correct answers due to the differentiation in human annotators. Hence, the model prediction is supervised by the occurrence probability $y_j \in [0, 1]$ of each candidate answer $A_j$ based on the human labeled answers.
%Specifically, UpDn adopts binary-cross-entropy loss to optimize the model parameters, 
%Hence UpDn utilizes the soft-scores $\boldsymbol{y}$ of the corresponding ground-truth answers $\hat{A}$ to supervise the model prediction. 

In fact, each question may have several correct answers due to the differentiation in human annotators. Hence, the model prediction is supervised by a set of soft values $\boldsymbol{y} \in {[0, 1]}^{\mathcal{|A|}}$, where ${y}_j$ denotes the occurrence probability of the candidate answer $A_j$ based on the human labeled answers.
Specifically, UpDn adopts the binary-cross-entropy loss to optimize the model parameters,
%\begin{equation}
%\label{loss_vqa}
%\mathcal{L}_{vqa} = \sum_{i=1} \boldsymbol{y}^{T}\log{(\boldsymbol{p})} - (1-\boldsymbol{y})^{T}\log{(1-\boldsymbol{p})}.
%\end{equation}

%\begin{equation}
%\label{loss_vqa}
%\mathcal{L} = \sum_{i} \sum_{j} {y}_{ij}\log{({p}_{ij})} - (1-{y}_{ij})\log{(1-{p}_{ij})},
%\end{equation}

\begin{equation}
\label{loss_vqa}
{
\mathcal{L} = -\sum_{i} \sum_{j} {y}_{ij}\log{({p}_{ij})} - (1-{y}_{ij})\log{(1-{p}_{ij})},
}
\end{equation}

\noindent { where} the indices $i$ and $j$ refer to the training questions and candidate answers, respectively. 
%\begin{equation}
%\label{loss_vqa}
%\mathcal{L}_{vqa} = \sum_{j=1}^{\mathcal{|A|}} {y}_j\log{({p}_j)} - (1-{y}_j)\log{(1-{p}_j)}.
%\end{equation}

\begin{algorithm}[t]
	\centering
	\caption{Typical Visual Attention-based Model (UpDn)}
	%	\textbf{Input:} Image $I$, Question $Q$, Ground-truth answer $\hat{A}$;\\
	%	\textbf{Output:} Binary-cross-entropy loss $\mathcal{L}$;\
	\label{alg:UpDn}	
	\begin{algorithmic}[1]
		\Function{{$\mathrm{UpDn}$}}{$I$, $Q$, $\hat{A}$}
%		\State $\mathcal{V}\gets {\rm RCNN}(I)$ \Comment{set of object features}
%		\State $\boldsymbol{q}\gets {\rm RNN}(Q)$ \Comment{question features}
		\State $\mathcal{V}\gets {\rm RCNN}(I), \quad\boldsymbol{q}\gets {\rm RNN}(Q)$ \Comment{image features and question features}
		\State $\boldsymbol{\alpha}\gets {\rm VAtt}(\mathcal{V}, \boldsymbol{q})$ \Comment{visual attention weights}
		\State $\hat{\boldsymbol{v}} \gets \sum_{i=1}^K {\alpha}_i \boldsymbol{v}_i$ \Comment{final visual representation}
		\State $\boldsymbol{p} \gets f_p(\hat{\boldsymbol{v}}, \boldsymbol{q})$ \Comment{predicted probabilities for all candidate answers}
		\State Compute soft values $\boldsymbol{y}$ for ground-truth answers $\hat{A}$
		\State Compute loss $\mathcal{L}$ according to Eqn. (\ref{loss_vqa})
		
		\State \Return $\boldsymbol{\alpha}$, $\mathcal{L}$
		\EndFunction
		
	\end{algorithmic}
\end{algorithm}

%			\State Image features $\boldsymbol{V}\gets {\rm RCNN}(I)$
%			\State Question features $\boldsymbol{q}\gets {\rm RNN}(Q)$
%			\State Attention weights $\boldsymbol{\alpha}\gets {\rm VAtt}(\boldsymbol{V}, \boldsymbol{q})$
%			\State Final visual representation $\hat{\boldsymbol{v}} \gets \sum_{i=1}^K {\alpha}_i \boldsymbol{v}_i$
%			\State Predicted probabilities $\boldsymbol{p} \gets f_p(\hat{\boldsymbol{v}}, \boldsymbol{q})$
%			\State Compute soft values $\boldsymbol{y}$ for $\hat{A}$
%			\State Compute loss $\mathcal{L}$ according to Eqn. \ref{loss_vqa}

%		$\lnot$ top-N\%($\boldsymbol{\alpha}$)
\begin{algorithm}[t]
	\caption{Visual Attention Regularization (AttReg)}
	\label{alg:AttReg}	
	%	\textbf{Input:} Image $I$, Question $Q$, Ground-truth Answer $\hat{A}$;\\
	%	\textbf{Output:} Predicted Answer $P$;\
	\begin{algorithmic}[1]
		\Function{{$\mathrm{AttReg}$}}{$I$, $Q$, $\hat{A}$} \Comment{{Backbone model (UpDn) finetuning}}
		\State $V^{*}\gets$ key objects identification 
		\State $\boldsymbol{\alpha}$, $\mathcal{L}_{vqa}\gets {\mathrm{UpDn}}(I, Q, \hat{A})$ \Comment{original training sample}
		\State $V^{o}\gets$ ignored objects localization
		
		\State $I^m\gets$  masking ignored key objects ($V^{*}\cap V^{o}$) in $I$ 
		%		\Comment{curated-image construction}
		\State $\hat{A}^m := \varnothing$ 
		\State \_, $\mathcal{L}_{reg}\gets {\mathrm{UpDn}}(I^m, Q, \hat{A}^m)$ \Comment{curated training sample}
		\State $\mathcal{L}_{all}\gets \mathcal{L}_{vqa} + \lambda \mathcal{L}_{reg}$ \Comment{parameter update}
		\EndFunction
	\end{algorithmic}
\end{algorithm}

\subsection{Visual Attention Regularization (AttReg)}
\label{approach_AttReg}
\begin{figure*}[t]
	\centering
	\includegraphics[width=0.99\textwidth]{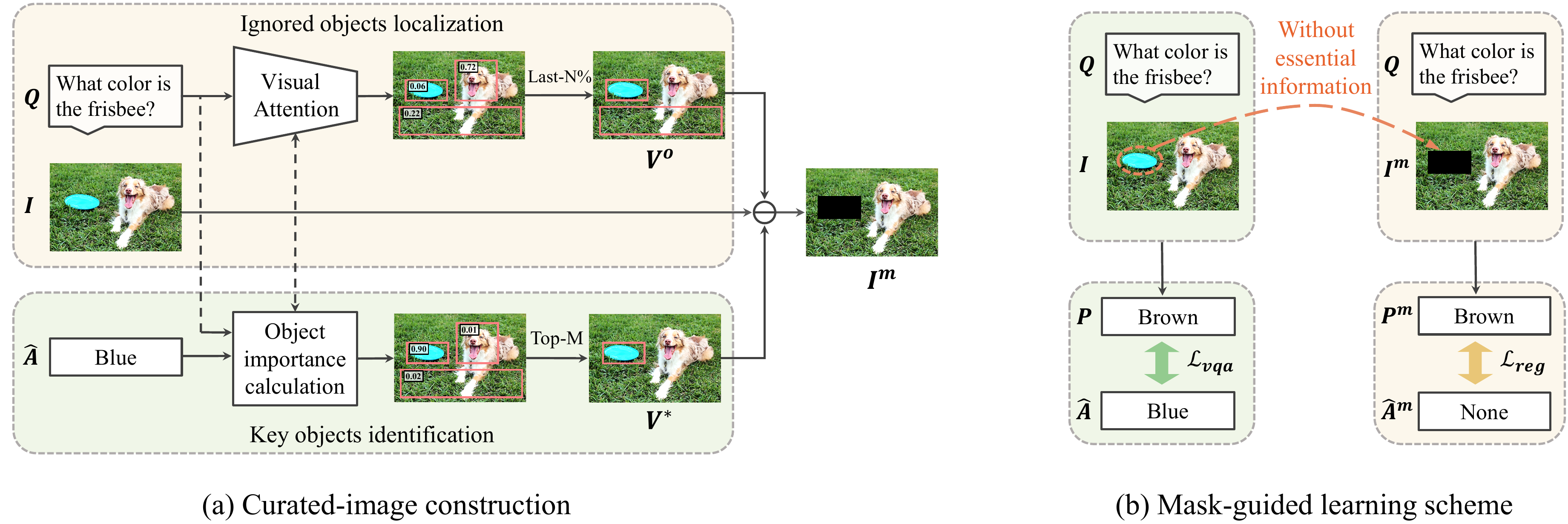} % Reduce the figure size so that it is slightly narrower than the column. Don't use precise values for figure width.This setup will avoid overfull boxes. 
	\caption{Illustration of the proposed AttReg for regularizing the visual attention. (a) Given a training sample, a curated-image is firstly constructed by masking the identified ignored key objects. (b) The AttReg then composes a new training sample to regularize the visual attention to focus more on the ignored key objects.}
	\label{model}
\end{figure*}

%Given a training sample $<Q, I, \hat{A}>$, we first identify the ignored important objects and construct a new curated image $I^m$ with these objects being masked (see Figure \ref{model1} for illustration). Then we append a complementary sample $<Q, I^m, \hat{A}^m>$ to train the model (see Figure \ref{model2} for illustration). In the following, we will sequentially introduce the detail of above two processes of AttReg.
Existing VQA models often make mistakes due to the inappropriate visual grounding. To tackle this issue, in this work, we regularize the visual attention for better model performance.
As shown in Figure \ref{model}, given a training sample $<Q, I, \hat{A}>$, we first identify the ignored key objects and construct a new curated image $I^m$ with these objects being masked. Then we append a complementary sample $<Q, I^m, \hat{A}^m>$ to regularize the model for better attention learning. 
{Overall, our method could be regarded as a data augmentation technique, which guides the model learning by synthesizing the curated training samples.
}
The following will sequentially introduce two procedures of AttReg.

\textbf{1. Curated-Image Construction}
The module of curated-image construction consists of three main steps:
(i) identifying key objects related to the QA pair, represented as $V^{*}$;
(ii) locating ignored objects through visual attention, represented as $V^{o}$; (iii) masking the ignored key objects (\textit{i.e.}, $V^{*}\cap V^{o}$) and constructing a new curated image ${I}^m$. 
%{\color{orange}}

Particularly, we firstly take the QA pair as the auxiliary information to determine the importance of each object in the image. Following \cite{Baseline_AttSupv:SCR}, we assign POS tags to each word in the QA using the spaCy POS tagger \cite{Tech:Spacy} and extract nouns in the QA.
Thereafter, we calculate the cosine similarity between the GloVe \cite{Tech:Glove} embedding of object categories and the extracted nouns.\footnote{Object categories are the outputs from the used faster-RCNN network, which can extract object features and predict its category simultaneously.}
{
And then, we select top-$M$ objects whose similarity scores are greater than threshold $\sigma$ as key objects $V^{*}$. 
}
%The top-$M$ objects with the highest similarity scores are selected as key objects $V^{*}$. 
We consider these objects as essential since they are highly related to the QA.
Note that directly leveraging these extracted key objects as the supervision is sub-optimal as studied in prior work \cite{Baseline_AttSupv(Argument):HINT}.
The dominant reason is that the presence of superficial linguistic correlations (\textit{i.e.}, language biases \cite{Dataset:VQA_v2,Baseline_LP:Visual_Bias}) can easily mislead the model to ignore visual content even under a strong supervision \cite{Baseline_AttSupv(Argument):HINT}. 
In the light of this, we instead apply these key objects to compose a new training sample, which can help the model build correlation between the right visual information and the ground-truth answer. The superficial bias can also be alleviated to some extent.

%\begin{figure*}[t]
%	\centering
%	\includegraphics[width=0.8\textwidth]{supplements/model2} % Reduce the figure size so that it is slightly narrower than the column. Don't use precise values for figure width.This setup will avoid overfull boxes. 
%	\caption{Illustration of the proposed AttReg for regularizing the visual attention. (a) Given a training sample, a curated-image is firstly constructed by masking the identified ignored key objects. (b) The AttReg then composes a new training sample to regularize the visual attention to focus more on the ignored key objects.}
%	\label{model2}
%\end{figure*}

%Refer to Figure \ref{att_table}, image objects with larger attention weights show higher more responsible for model decisions.
%due to the superior faithfulness of visual attention, 

Secondly, we take the attention weights $\boldsymbol{\alpha}$ as the evaluation standard for examining the influence of each object {in model prediction}. 
Specifically, to avoid trivial procedures, we directly {select image objects with attention weight rank in the {last-$N$\%} as the ignored objects $V^{o}$}.
{The main idea is that remaining merely a few image objects with the highest attention weights can achieve very competitive performance (\textit{e.g.}, a minor drop of -0.55\% if only remaining top-20\%), which implies that most objects ranked lower hardly impose an effect on model predictions; as a result, we can deem such objects as ignored by the model.}
%{Note that while many works employ Grad-CAM for visual grounding, we can see that it is not faithful to model predictions, because a model using features with low Grad-CAM weights even outperforms the one using larger weights (Table \ref{att_faith_table}). Hence visual attention is more suitable to quantify the influence of an object (see Section \ref{Sec_VAttExp} for more details).}

%Secondly, we take the visual attention weights $\boldsymbol{\alpha}$ as the evaluation standard for examining the influence of each object towards model decisions. 
%{Refer to Table \ref{att_faith_table}}, remaining merely $N$\% (\textit{e.g.}, 20\%) image objects with the highest attention weights can achieve very competitive performance (63.42\% vs. 63.97\%), which implies that most image objects out of top-$N$\% hardly impose an effect on the model predictions. 
%Consequently, to avoid trivial procedures, we directly select image objects that are out of top-$N$\% attention weight rank as the ignored objects $V^{o}$. 

Lastly, we use the intersection of $V^{*}$ and $V^{o}$ as the ignored key objects and mask them in the original image ${I}$ to construct a new curated image ${I}^m$. 
{
It should be mentioned that the above preprocessing steps are consistent with existing studies  \cite{AttentionMethods:UpDn,Baseline_AttSupv:CSS,Baseline_AttSupv:SCR,Baseline_LP:CL,Baseline_LP:Decomp-LR,Baseline_LP:RuBi,Baseline_LP:Sup_2020,R2:OOD,R1:negative_analysis,Baseline_LP:LMH,AttentionMethods:EraseAtt,Baseline_LP:AdvReg,adaVQA}, \textit{e.g.}, employing the same faster-RCNN features and POS Tagger. Thus, while inevitable limitations might exist, we choose to neglect them following prior works for a fair comparison. 
}

\begin{figure*}[t]
	\centering
	\includegraphics[width=0.85\textwidth]{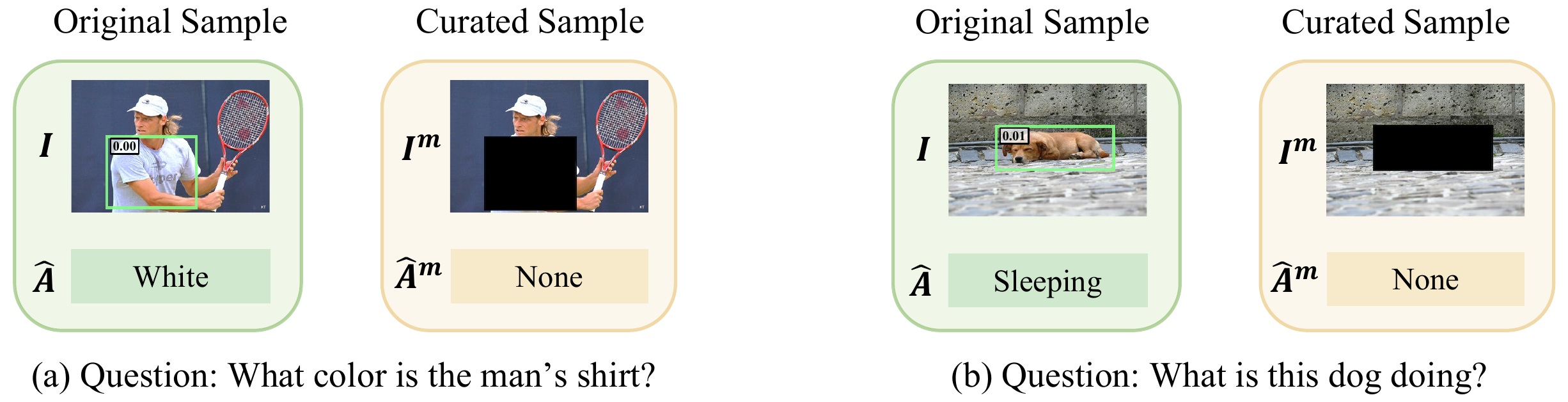} % Reduce the figure size so that it is slightly narrower than the column. Don't use precise values for figure width.This setup will avoid overfull boxes. 
	\caption{Two running examples from AttReg. The green boxes denote the ignored key objects, and the value around the bounding box is the attention weight to the given object.}
	\label{samples}
\end{figure*}

\textbf{2. Mask-guided Learning Scheme}
Based on the curated image $I^m$, we then design a mask-guided learning scheme to regularize the model, which emphasizes more on the ignored key objects towards question answering. 
We argue that a VQA model cannot correctly answer the question if it is blind to the key objects from the given image. In a sense, the curated image $I^m$ lacks essential visual information in correctly answering the question $Q$ because the key objects are masked. 
We hence simply zero-out all ground-truth answers in $\hat{A}$ for $<Q, I^m>$ and compose a new training sample $<Q, I^m, \hat{A}^m>$, \textit{i.e.}, $\hat{A}^m$ is $\varnothing$ {(more examples can be found in Figure \ref{samples})}. 
{
In this way, when the training image loses the ignored key objects, the model would be trained towards a different target; thereby AttReg can constrain such ignored objects to yield stronger influence in model prediction. More importantly, this influence can be seamlessly transferred to the attention weights learning because of their positive correlations (Figure \ref{att_faith_fig} (b)).
}

%In this way, the devised mask-guided learning scheme enforces the variation in the target answer distribution when the training image loses the ignored key objects, thereby constraining these ignored objects to yield higher influence towards model predictions. More importantly, this influence can be seamlessly transferred to the attention weights learning since these two are positively correlated (refer to Figure \ref{att_faith_fig} (b)). As such, the attention weight improvement of ignored key objects can be eventually achieved.

On the other hand, by training the model with $<Q, I^m, \hat{A}^m>$ and $<Q, I, \hat{A}>$ simultaneously, the existence of the masked visual information (\textit{i.e.}, ignored key objects) would be strengthened as a necessary condition for predicting the correct answers, as illustrated in Figure \ref{model} (b). Hence, the strong correlation between the ignored key objects and the ground-truth answers can be built under this situation.
%The training algorithm of our method is summarized at Algorithm \ref{alg:AttReg}.
{ We summarize this method in Algorithm \ref{alg:AttReg}.}

%In this way, the devised mask-guided learning scheme enforces the variation in target answer distribution when the training image loses the ignored key objects, regularizing the model to `see' the ignored visual information as well as assign higher attention weights on these objects. The training algorithm of our method can be found at XX.
%In practice, this design is inspired by the faithfulness of visual attention -- \textit{the attention weight value is highly correlated with the contribution of the image region to model predictions}. 
%By training the model with $<Q, I^m, \hat{A}^m>$ and $<Q, I, \hat{A}>$ simultaneously, we enforces the variation in target answer distribution when the training image loses the ignored key objects, which can leads to the enhancement in the influence of these ignored objects towards answer predictions. 
%This is to say, the attention weight improvement of ignored objects can be eventually achieved by such influence enhancement.

Overall, we optimize the parameters in our model by simultaneously training with the original and curated samples,
\begin{equation}
\label{all_loss}
\mathcal{L}_{all} = \mathcal{L}_{vqa} + \lambda \mathcal{L}_{reg},
\end{equation}
where $\mathcal{L}_{vqa}$ and $\mathcal{L}_{reg}$ respectively denote the original sample training loss and curated sample training loss, and $\lambda$ is a hyper-parameter to control the regularization strength. 
{
In the training phase, the generation of a curated sample is determined by the number of ignored key objects. Specifically, if the number of the ignored key objects is $\geq$1, one sample will be curated for model learning, otherwise not.
}
In the testing phase, the regularization module is no longer activated and only the backbone model remains.

\section{Experiments}
\label{Sec_exp}
%We conduct extensive experiments on two popular datasets to thoroughly justify the effectiveness of our proposed method
\subsection{Datasets and Experimental Settings}
\bfstart{Datasets}
We evaluated the proposed AttReg mainly on the diagnostic VQA-CP datasets \cite{Dataset:VQA_CP(GVQA)}, wherein the answer distributions for each question category in the training and testing sets are significantly different. As such, most VQA models that lack visual grounding ability and rely on superficial correlations between questions and answers (\textit{i.e.}, language biases) would perform poorly on this dataset.
%In addition, since the VQA-CP is a re-organized version of the VQA v2 \cite{Dataset:VQA_v2}, we also reported the experimental results on the VQA v2 for completeness.
The experimental results on the VQA v2 \cite{Dataset:VQA_v2} are also reported for completeness.

\bfstart{Evaluation Metrics}
We adopted the standard VQA accuracy metric \cite{Dataset:VQA_v1} to evaluate the model performance, which is defined as,
\begin{equation}
\label{acc_metric}
ACC(ans) = \min\left\{1, \frac{\mbox{\#humans\ that\ provide\ } ans}{3}\right\}.
\end{equation}
Note that each question is answered by ten annotators, and ACC considers the disagreement among human answers. 

\bfstart{Implementation and Training Details}
We implemented our method following the preprocessing steps of the {widely adopted} UpDn  \cite{AttentionMethods:UpDn} as well as the LMH  \cite{Baseline_LP:LMH} for a fair comparison. 
Specifically, for each image, the UpDn utilizes Faster-RCNN to generate 36 object proposals as the visual features, \textit{i.e.}, a 2,048-d vector for each object. 
All the questions are converted to lower case and trimmed to a maximum of 14 words. And the questions with less than 14 words are padded with zeros. The pre-trained GloVe vectors are used to initialize the word embedding matrix, \textit{i.e.}, a 300-d vector for each word. And then a single-layer GRU is employed to obtain sentence-level question encoding of 512-d.

We first pretrained the backbone model on the training splits using the plain VQA loss $\mathcal{L}_{vqa}$. We then regularized the visual attention for the backbone model by fine-tuning with $\mathcal{L}_{all}$ in Eqn. (\ref{all_loss}).
For UpDn and LMH, the learning rate is set to 2e-3 during pre-training, and reduced to 2e-5 and 2e-4 when fine-tuning, respectively.
{In addition, the $\lambda$ of AttReg is set to 1.0 and 0.5 for the UpDn and LMH, respectively.
And for the ignored objects, the proportion of $N$\% is tuned between 10\% and 40\%. The number of the key objects $M$ is tuned between 1 and 6. }

\begin{table*}[t]
	\caption{Performance comparison between the proposed method and the state-of-the-arts on VQA-CP v2 test and VQA v2 val. \textit{Expl.} implies the source of the explanations, \textit{e.g.}, human attention (HAT) \cite{Dataset:HAT}. \textit{Att.} represents the visual attention regularization approach. \textit{Grad.} stands for the Grad-CAM methods. $^\dagger$ denotes our implementation. The mean score represents the accuracy average on VQA-CP v2 and VQA v2.}\smallskip
	\centering
	\resizebox{0.99\textwidth}{!}{
		\smallskip\begin{tabular}{l c c c  c c c c  c c c c  c}
			\toprule[1pt]
			%			SCR \cite{Baseline_AttSupv:SCR} & VQA-X & & \checkmark & 49.45 & 72.36 & 10.93 & 48.02 & 62.20 & 78.80 & 41.60 & 54.50 \\
			\multirow{2}{*}{Method} &\multirow{2}{*}{Expl.} &\multirow{2}{*}{Att.} &\multirow{2}{*}{Grad.} & \multicolumn{4}{c}{VQA-CP v2 test} & \multicolumn{4}{c}{VQA v2 val} & Mean\\
			\cmidrule(lr){5-8} \cmidrule(lr){9-12} \cmidrule(l){13-13}
			& & & & All & Yes/No & Number & Other & All & Yes/No & Number & Other & All \\ \midrule
			
			SAN (2016) \cite{AttentionMethods:SAN} & & & & 24.96 & 38.35 & 11.14 & 21.74 & 52.41 & 70.06 & 39.28 & 47.84 & 38.69\\
			NMN (2016) \cite{Baseline:NMN} & & & & 27.47 & 38.94 & 11.92 & 25.72 & 51.62 & 73.38 & 33.23 & 39.93 & 39.55 \\
			HAN (2018) \cite{AttentionMethods:HardAtt} & & & & 28.65 & 52.25 & 13.79 & 20.33 & - & - & - & - & -\\
			GVQA (2018) \cite{Dataset:VQA_CP(GVQA)} & & & & 31.30 & 57.99 & 13.68 & 22.14 & 48.24 & 72.03 & 31.17 & 34.65 & 39.77 \\
			MCB (2016) \cite{AttentionMethods:MCB} & & & & 36.33 & 41.01 & 11.96 & 40.57 & 59.71 & 77.91 & 37.47 & 51.76 & 48.02 \\
			
			\midrule

			MuRel (2019) \cite{Baseline:MuRel} & & & & 39.54 & 42.85 & 13.17 & 45.04 & - & - & - & - & - \\
			AdvReg (2018) \cite{Baseline_LP:AdvReg} & & & & 41.17 & 65.49 & 15.48 & 35.48 & 62.75 & 79.84 & 42.35 & 55.16 & 51.96 \\
			RUBi (2019) \cite{Baseline_LP:RuBi} & & & & 47.11 & 68.65 & 20.28 & 43.18 & 61.16 & - & - & - & 54.14 \\
			EraseAtt (2019) \cite{AttentionMethods:EraseAtt} & & & & 37.43 & 41.98 & 13.00 & 41.74 & 65.99 & - & - & - & 51.71 \\
			DecompLR (2020) \cite{Baseline_LP:Decomp-LR} & & & & 48.87 & 70.99 & 18.72 & 45.57 & 57.96 & 76.82 & 39.33 & 48.54 & 53.42 \\
			SimpleReg (2020) \cite{R1:negative_analysis} & & & & 48.90 & - & - & - & 62.60 & - & - & - & 55.75 \\
			
			RMFE (2020) \cite{Baseline_LP:RMFE} & & & & 54.55 & 74.03 & 49.16 & 45.82 & - & - & - & - & - \\
			RandImg (2020) \cite{R2:OOD} & & & & 55.37 & 83.89 & 41.60 & 44.20 & 57.24 & 76.53 & 33.87 & 48.57 & 56.31 \\
			SelfSup (2020) \cite{Baseline_LP:Sup_2020} & & & & 57.59 & 86.53 & 29.87 & \textbf{50.03} & 63.73 & - & - & - & 59.55 \\
			CL (2020) \cite{Baseline_LP:CL} & & & & 59.18 & 86.99 & 49.89 & 47.16 & 57.29 & 67.27 & 38.40 & 54.71 & 58.24 \\
			\midrule
			UpDn (2018) \cite{AttentionMethods:UpDn} & & & & 39.74 & 42.27 & 11.93 & 46.05 & 63.48 & 81.18 & 42.14 & 55.66 & 51.61 \\
			UpDn$^\dagger$ (2018) \cite{AttentionMethods:UpDn} & & & & 40.09 & 42.16 & 12.36 & 46.61 & 63.77  & 81.54 & 43.64 & 55.59 & 51.93 
			\\
			UpDn-SCR (2019) \cite{Baseline_AttSupv:SCR} & QA & & \checkmark & 48.47 & 70.41 & 10.42 & 47.29 & 62.30 & 77.40 & 40.90 & \textbf{56.50} & 55.39 \\
			UpDn-HINT (2019) \cite{Baseline_AttSupv(Argument):HINT} & HAT & & \checkmark & 46.73 & 67.27 & 10.61 & 45.88 & 63.38 & 81.18 & 42.99 & 55.56 & 55.06 \\
			UpDn-AttAlign (2019) \cite{Baseline_AttSupv(Argument):HINT}  & HAT & \checkmark & & 39.37 & 43.02 & 11.89 & 45.00 & 63.24 & 80.99 & 42.55 & 55.22 & 51.31 \\
			UpDn-CSS (2020) \cite{Baseline_AttSupv:CSS} & QA & & \checkmark & 41.16 & 43.96 & 12.78 & 47.48 & - & - & - & - & - \\
			%			UpDn-AttReg (Ours) & QA & \checkmark & & 46.85 & 69.34 & 12.44 & 44.51 & \textbf{64.13} & \textbf{81.72} & \textbf{43.77} & 56.13 & 55.49 \\
			UpDn-AttReg (Ours) & QA & \checkmark & & 46.75 & 66.23 & 11.94 & 46.09 & \textbf{64.13} & \textbf{81.72} & \textbf{43.77} & 56.13 & 55.44 \\
			
			\midrule
			%			LMH-CSS \cite{Baseline_AttSupv:CSS} & QA & 58.95 & 84.37 & 49.42 & \textbf{48.21} & 59.91 & 73.25 & 39.77 & 55.11 & \\
			LMH (2019) \cite{Baseline_LP:LMH}  & & & & 52.45 & 69.81 & 44.46 & 45.54 & 61.64 & 77.85 & 40.03 & 55.04 & 57.05 \\
			LMH$^\dagger$ (2019) \cite{Baseline_LP:LMH} & & & & 52.99 & 72.02 & 39.24& 46.79 & 62.40 & 79.42 & 41.48 & 54.99 & 57.70 \\
			LMH-SCR (2019) \cite{Baseline_AttSupv:SCR} & QA & & \checkmark & 52.31 & 73.43 & 36.75 & 45.51 & - & - & - & - & - \\
			LMH-HINT (2019) \cite{Baseline_AttSupv(Argument):HINT} & HAT & & \checkmark & 52.58 & 74.41 & 36.78 & 45.48 & - & - & - & - & -\\
			LMH-CSS (2020) \cite{Baseline_AttSupv:CSS} & QA & & \checkmark & 58.95 & 84.37 & 49.42 & 48.21 & 59.91 & 73.25 & 39.77 & 55.11 & 59.43\\
			%			LMH-AttReg (Ours)  & QA & \checkmark & & \textbf{59.92} & \textbf{87.28} & \textbf{52.39} & 47.65 & 62.74 & 79.71 & 41.68 & 55.42 & \textbf{61.33} \\
			LMH-AttReg (Ours)  & QA & \checkmark & & \textbf{60.00} & \textbf{86.80} & \textbf{51.63} & 48.25 & 62.74 & 79.71 & 41.68 & 55.42 & \textbf{61.37} \\

			\bottomrule[1pt]
		\end{tabular}
	}
	\label{acc_total}
\end{table*}

\subsection{Comparisons with State-of-the-Arts}
\bfstart{Performance on VQA-CP v2 and VQA v2}
Table \ref{acc_total} exhibits the results of our method and the SOTA VQA models on VQA-CP v2 and VQA v2. 
%{
%The first part in this table exhibits plain VQA models, followed by the models build upon the UpDn or LMH. And the last two parts denote the backbone models with or without visual grounding enhancement methods, \textit{i.e.}, SCR \cite{Baseline_AttSupv:SCR}, HINT \cite{Baseline_AttSupv(Argument):HINT}, AttAlign \cite{Baseline_AttSupv(Argument):HINT}, CSS \cite{Baseline_AttSupv:CSS}, and our AttReg. 
%Further, SCR and HINT were also equipped into the well-devised LMH model for fair comparison, and the results are reported in Table \ref{acc_lmh_compare}. }
{The test results of our method on VQA v2 are illustrated in Table \ref{VQAv2_test}.}
The main observations are as follows.

Firstly, our LMH-AttReg achieves a new state-of-the-art of {60.00\%} ACC on the VQA-CP v2 dataset. {Specifically, the LMH-AttReg performs better than the existing best approach CL over all three answer categories. In addition, our LMH-AttReg also achieves the highest mean score over the two datasets among the existing state-of-the-arts, \textit{i.e.}, LMH-AttReg vs. SelfSup ({61.37\%} vs. 59.55\%)}

Secondly, our AttReg dramatically improves the backbone model performance on the VQA-CP v2 dataset. By incorporating AttReg into UpDn, a large improvement is achieved (\textit{i.e.}, {+6.66\%}). And LMH-AttReg also yields a large gain (\textit{i.e.}, {+7.01\%}) over its backbone LMH, which highlights the effectiveness and the generalization capability of AttReg.

\begin{table*}[t]
	\centering
	\caption{Performance of the proposed method on VQA v2 test-std. $^\dagger$ denotes our implementation.}
	\resizebox{0.55\columnwidth}{!}{
		\smallskip\begin{tabular}{l c c c c}
			\toprule[1pt]
			
			Method & All & Yes/No & Number & Other
			\\ \midrule
			UpDn$^\dagger$  & 64.90 & 81.82 & 42.73 & 55.51 \\
			UpDn-AttReg (Ours) & \textbf{65.25} & \textbf{82.18} & \textbf{43.68} & \textbf{55.71} \\
			
			\midrule 
			LMH$^\dagger$  & 63.31 & 78.15 & \textbf{41.08} & 55.48 \\
			LMH-AttReg (Ours) & \textbf{63.43} & \textbf{78.28} & 40.7 & \textbf{55.72} \\
			
			\bottomrule[1pt]

		\end{tabular}
	}
	\label{VQAv2_test}
\end{table*}

\begin{table*}[t]
	\caption{Performance comparison between the proposed method and state-of-the-arts on VQA-CP v1 test.}\smallskip
	\centering
	\resizebox{0.55\columnwidth}{!}{
		\smallskip\begin{tabular}{l c c c c }
			\toprule[1pt]
			
			Method & All & Yes/No & Number & Other
			\\ \midrule
			SAN \cite{AttentionMethods:SAN} & 26.88 & 35.34 & 11.34 & 24.70\\
			NMN \cite{Baseline:NMN} & 29.64 & 38.85 & 11.23 & 27.88\\
			GVQA \cite{Dataset:VQA_CP(GVQA)} & 39.23 & 64.72 & 11.87 & 24.86\\
			MCB \cite{AttentionMethods:MCB} & 34.39 & 37.96 & 11.80 & 39.90\\
			AdvReg \cite{Baseline_LP:AdvReg} & 43.43 & 74.16 & 12.44 & 25.32\\
			RUBi \cite{Baseline_LP:RuBi} & 50.90 & 80.83 & 13.84 & 36.02\\
			CL \cite{Baseline_LP:CL} & 61.27 & 88.14 & 34.43 & 45.34\\
			%			AdaVQA \cite{adaVQA} & 61.20 & \textbf{91.17} & 41.34 & 39.38\\
			\midrule
			UpDn$^\dagger$ \cite{AttentionMethods:UpDn} & 38.88 & 42.48 & 13.12 & 45.66\\
			LMH$^\dagger$ \cite{Baseline_LP:LMH} & 55.73 & 78.59 & 24.68 & 45.47\\
			LMH-CSS \cite{Baseline_AttSupv:CSS} & 60.95 & 85.60 & \textbf{40.57} & 44.62\\
			UpDn-AttReg (Ours) & 47.66 & 66.17 & 12.11 & 43.54\\
			LMH-AttReg (Ours) & \textbf{62.25} & \textbf{88.29} & 35.30 & \textbf{47.18}\\
			
			\bottomrule[1pt]
		\end{tabular}
		
	}
	\label{acc_cpv1}
\end{table*}

%, \textit{i.e.}, -0.38\% for UpDn and -0.10\% for LMH
Thirdly, our AttReg can enhance the backbone model performance on the VQA v2 dataset over all answer categories. In contrast, most approaches that perform well on VQA-CP v2 suffer performance drop over this dataset. For example, introducing SCR \cite{Baseline_AttSupv:SCR} into UpDn leads to a significant improvement of +8.65\% on VQA-CP v2; whereas the 1.47\% performance drop can be observed when SCR works on VQA v2. By comparison, AttReg can enhance UpDn on both VQA-CP v2 ({+6.66\%}) and VQA v2 (+0.36\%), which further reveals the robustness of AttReg.

Fourthly, we compared our AttReg with a straightforward visual attention regularization method -- AttAlign \cite{Baseline_AttSupv(Argument):HINT}, which directly aligns visual attention with human attentions. The result demonstrates that our AttReg outperforms AttAlign with a large margin on the two datasets, especially on VQA-CP v2 ({46.75\%} vs. 39.37\%). This proves the effectiveness and superiority of our method among the visual attention regularization ones.

{Lastly, we compared our AttReg with the existing gradient-based regularization methods.}
It can be found that while these gradient-based methods enhance model performance significantly on VQA-CP v2, they usually deteriorate the backbone model performance on VQA v2. 
Instead, our AttReg is able to not only improve the model performance on the two datasets, but also achieve the highest mean score over the two datasets among the existing grounding enhancement approaches, \textit{i.e.}, LMH-AttReg vs. LMH-CSS ({61.37\%} vs. 59.43\%) and UpDn-AttReg vs. UpDn-SCR ({55.44\%} vs. 55.39\%). 
{
	Furthermore, when incorporating SCR and HINT into the LMH backbone, the model performance even exhibits a decrease (similar experimental results can be found in \cite{Baseline_AttSupv:CSS}). In contrast, our AttReg performs well on both backbones.
}

%This further proves our previous finding (in Section \ref{Sec_VAttExp}): visual attention is more faithful to model decisions. In this way, a favorable visual attention regularization approach can yield more robust enhancement over the backbone model.

%i) typical VQA models; ii) models designed to overcome superficial biases; and iii) the backbone models with and without visual grounding enhancement methods, \textit{i.e.}, CSS and our AttReg. 
\bfstart{Performance on VQA-CP v1}
Table \ref{acc_cpv1} shows the results of our method and the SOTA VQA models on the VQA-CP v1 dataset. 
%Similarly, we group these models into: 
%{i) plain VQA models; ii) models builds upon the UpDn or LMH; and iii) the backbone models with and without visual grounding enhancement methods, \textit{i.e.}, CSS and our AttReg.
%}
From the results, we can observe that our LMH-AttReg achieves a new state-of-the-art performance on VQA-CP v1. 
In addition, our AttReg significantly improves the backbone performance on VQA-CP v1, \textit{i.e.}, +6.52\% on LMH and +8.78\% on UpDn.

\begin{table*}[t]
	\caption{In-domain and out-of-distribution (OOD) comparison between the proposed method and the baselines. 
		The first group denotes plain approaches, followed by the methods applied on UpDn.  $^\dagger$ denotes our implementation.
		As suggested in \cite{R2:OOD}, the results are selected with two requirements: (1) only using the val set for selection (without peeking at test data); (2) models share the same settings as the ones in Table \ref{acc_total}. 
	}\smallskip
	\centering
	\resizebox{0.95\columnwidth}{!}{
		\smallskip\begin{tabular}{l c c c c c c c c c}
			\toprule[1pt]
			\multirow{2}{*}{Method} & \multicolumn{4}{c}{VQA-CP v2 val (in-domain)} & \multicolumn{4}{c}{VQA-CP v2 test (OOD)} & Mean\\
			\cmidrule(lr){2-5} \cmidrule(lr){6-9}  \cmidrule(lr){10-10}
			& All & Yes/No & Number & Other & All & Yes/No & Number & Other & Other \\ \midrule
			
			Random-predictions & 37.62 & 70.10 & 32.79 & 10.55 & 10.44 & 25.87 & 9.27 & 2.57 & 6.56\\
			Random-predictions-inverted & 24.35 & 55.36 & 11.12 & 0.00 & 31.81 & 83.25 & 49.30 & 0.02 & 0.01\\
			\midrule
			%			UpDn & 64.73 &79.45	&49.59	&55.66		&38.82	&42.98	&12.18	&43.95		&51.78\\
			UpDn$^\dagger$ & 67.91 &83.82	&49.86	&57.97		&39.83	&43.2	&12.36	&45.61		&51.79\\
			
			%			UpDn & \textbf{68.1} & 40.0 & 54.05\\
			%			\cmidrule{2-4}
			%			
			UpDn-TopAnsMasked & 30.96 & 44.12 & 25.00 & 20.85 & 40.61 & 82.44 & 27.63 & 22.26 & 21.56\\
			UpDn-RandImg & 54.24 & 64.22 & 34.40 & 50.46 & \textbf{55.37} & \textbf{83.89} & \textbf{41.60} & 44.20 & 47.33\\
			UpDn-SCR & 67.09 &	82.57	&48.33	&57.73		&42.76	&53.02	&12.24	&45.77		&51.75\\
			UpDn-HINT & 67.2	&82.86	&46.13	&58.27		&42.5	&51.43	&11.76	&46.25		&52.26
			\\
			%			UpDn-AttReg (Ours) & 67.34	&83.27	&49.08	&57.42		&42.86	&51.64	&12.81	&\textbf{46.51}		&\textbf{55.1}
			UpDn-AttReg (Ours) & \textbf{68.23}	&\textbf{84.49}	&\textbf{49.87}	&\textbf{58.28}		&41.12	&45.01	&12.91	&\textbf{46.82}		&\textbf{52.55}
			\\
			%			
			%			\midrule
			%			LMH & 64.25	&77.63	&44.25	&57.14		&49.54	&72.5	&16.32	&46.62		&56.90
			%			\\
			%			
			%			\cmidrule{2-4}
			%			LMH-SCR & 64.66	&77.86	&45.22	&\textbf{57.88}		&49.76	&73.08	&16.02	&46.79		&57.21
			%			\\
			%			LMH-HINT & \textbf{65.11}	&\textbf{78.61}	&\textbf{45.3}	&57.85		&49.79	&73.87	&15.38	&46.61		&57.45
			%			\\
			%			LMH-AttReg & 62.92	&74.76	&42.19	&57.42		&\textbf{52.05}	&\textbf{77.29}	&\textbf{21.79}	&\textbf{47.14}		&\textbf{57.49}
			%			\\
			
			\bottomrule[1pt]
		\end{tabular}
		
	}
	\label{OOD:cpv2_val}
\end{table*}

{
\subsection{In-Domain and Out-of-Domain Testing}
\bfstart{Settings}
Following \cite{R2:OOD}, we randomly held out 8,000 instances from the VQA-CP v2 train set to {measure the} in-domain performance (which we will refer to as ``val'' set), and we utilized the VQA-CP v2 test set to measure the out-of-domain (OOD) performance.
{In addition,} as reported in \cite{R1:negative_analysis,R2:OOD}, by exploiting the difference in the answer distribution, many ordinary baselines could achieve very high performance on the VQA-CP v2 {test set}, especially on the ``Yes/No'' and ``Number'' answer categories. 
{Thus, we further added these baselines in the following and compared them with our AttReg on both the in-domain and OOD sets.}
%Thus, we further compared our AttReg with these baselines on both the in-domain and OOD sets.

\textit{Random-predictions.} It samples answers at random according to the distribution of answers observed per individual question type in the training set.

\textit{Random-predictions-inverted.} This method exploits the knowledge that the distribution over
the test data is approximately proportional to the inverse of the distribution over the training
data.

\textit{TopAnsMasked.} It assigns, at test time, the lowest possible score to the answer of highest predicted score.

\textit{RandImg.} This method augments the training data $<Q, V, \hat{A}>$ with a copy of the same question paired with random image features $V^*$: $<Q, V^*, \hat{A}>$. Moreover, in the training phase, it encourages the score of the correct answer to be low if the given image is $V^*$.

\bfstart{Comparison with Ordinary Baselines}
Table \ref{OOD:cpv2_val} exhibits the performance of our methods and the baselines across the in-domain and OOD sets. From the results, we could observe that 1) our UpDn-AttReg achieves the best mean score {(on the ``Other'' category)} over the two datasets among four ordinary baselines, \textit{e.g.}, UpDn-AttReg {(52.55\%)} vs. RandImg {(47.33\%)}; 2) RandImg and TopAnsMasked demonstrate great improvements over the ``Yes/No'' and ``Number'' categories on the OOD. However, their in-domain performance drops a lot on the same categories, which is exactly a clue that their obtained improvements are achieved by exploiting the difference in answer distributions. 
{
In contrast, our AttReg is hardly affected by this issue, and improves model performance over all answer categories on two datasets.
}
%In contrast, our AttReg can greatly alleviate this issue. Specifically, in the ``Yes/No'' category, our AttReg only deteriorates the in-domain performance with a very minor drop of -0.55\%, and the improvement on the OOD is +8.41\%. 

%trained on the subset of VQA-CP v2 training (\textit{i.e.}. exclude 8000 instances for val); 

\bfstart{Comparison with SCR and HINT} 
%To further validate the effectiveness of AttReg, we measured its in-domain and OOD performance on VQA-CP v2 and compared it with SCR and HINT (see Table \ref{OOD:cpv2_val}). 
To further validate the effectiveness of AttReg, {we compared its performance with SCR and HINT on the in-domain and OOD datasets (see Table \ref{OOD:cpv2_val}).
It can be found that, while the SCR and HINT improve model performance on the OOD set, their accuracy on the in-domain set exhibits a decrease compared with the baseline UpDn. 
%This behavior conforms with the newly added ordinary baselines (\eg, TopAnsMasked) – improving test performance by exploiting the distribution shift.
Contrary to that, our AttReg improves model performance on both sets, and achieves the highest mean score on the ``Other'' category, demonstrating advantages of AttReg over these two methods. 
}
%It can be found that, while the AttReg leads to a minor drop of UpDn on the val set, it can achieve a higher accuracy than both the SCR and HINT on both sets, demonstrating our AttReg's advantage over these two methods. 

%One possible reason for this drop is that AttReg may also play a role of alleviate over-fitting because many curated samples are synthesized during the training stage to guide model learning.

\begin{figure*}
	[t]
	\centering
	\includegraphics[width=0.8\columnwidth]{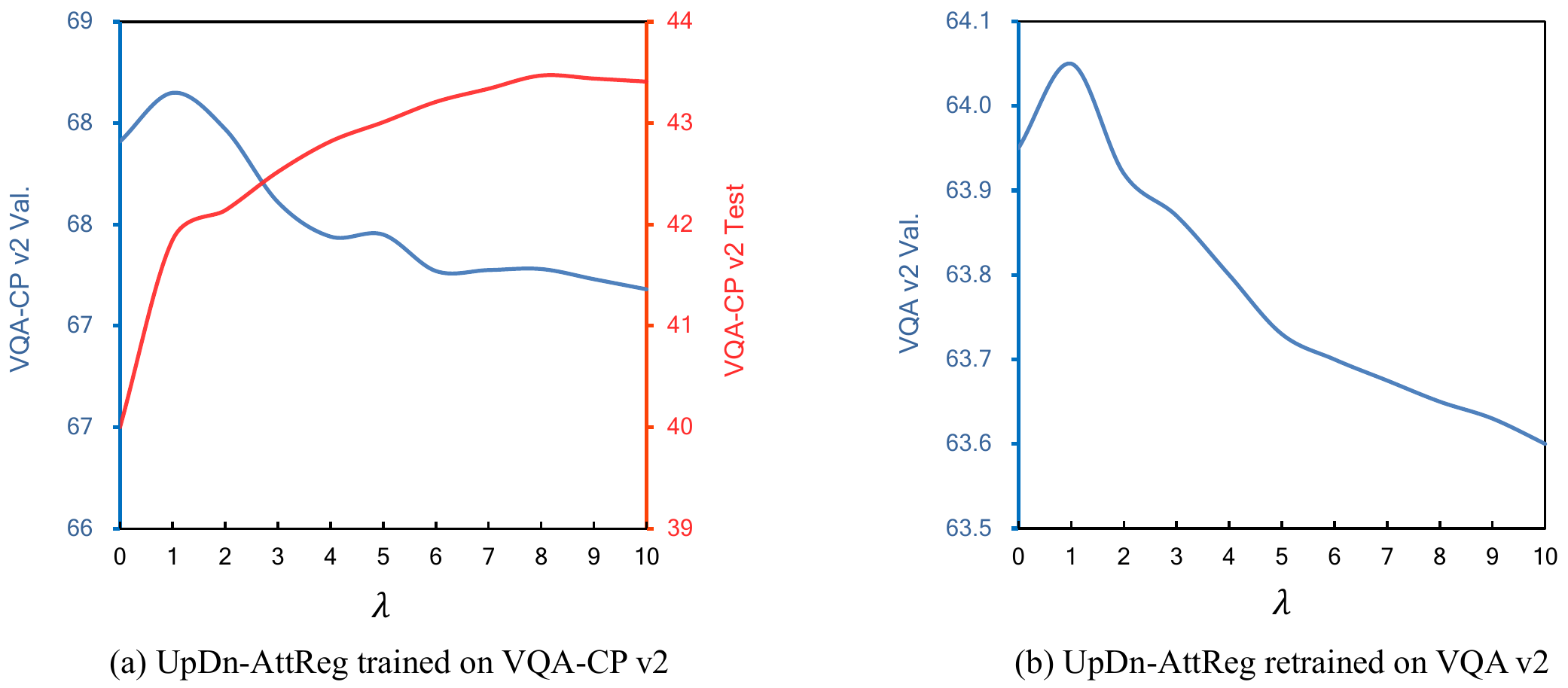} % Reduce the figure size so that it is slightly narrower than the column. Don't use precise values for figure width.This setup will avoid overfull boxes. 
	\caption{Performance change curves regarding the regularization strength $\lambda$ over VQA-CP v2 and VQA v2. The model retrained on VQA v2 shares the same setting as the corresponding model in VQA-CP v2.
	}
	\label{OOD:cpv2_updn_lambda}
	
\end{figure*}

\bfstart{Retraining Results}
As suggested in \cite{R2:OOD}, the regularization weight $\lambda$ tunes the trade-off between in-domain and OOD performance. As such, to further study the function of $\lambda$, we plotted the performance of {UpDn-AttReg} in Figure \ref{OOD:cpv2_updn_lambda}. It can be observed that 1) in most cases on the VQA-CP v2 dataset, a higher $\lambda$ can help improve the model performance on the OOD test set, and simultaneously the in-domain set performance drops; 2) when adopting an appropriate $\lambda$ (\textit{e.g.}, $\lambda$=1 for UpDn-AttReg), AttReg improves model performance on the all sets, \textit{i.e.}, in-domain and OOD sets in VQA-CP v2, and VQA v2.
}

\subsection{Ablative Studies}

\begin{figure*}
	[t]
	\centering
	\includegraphics[width=0.8\columnwidth]{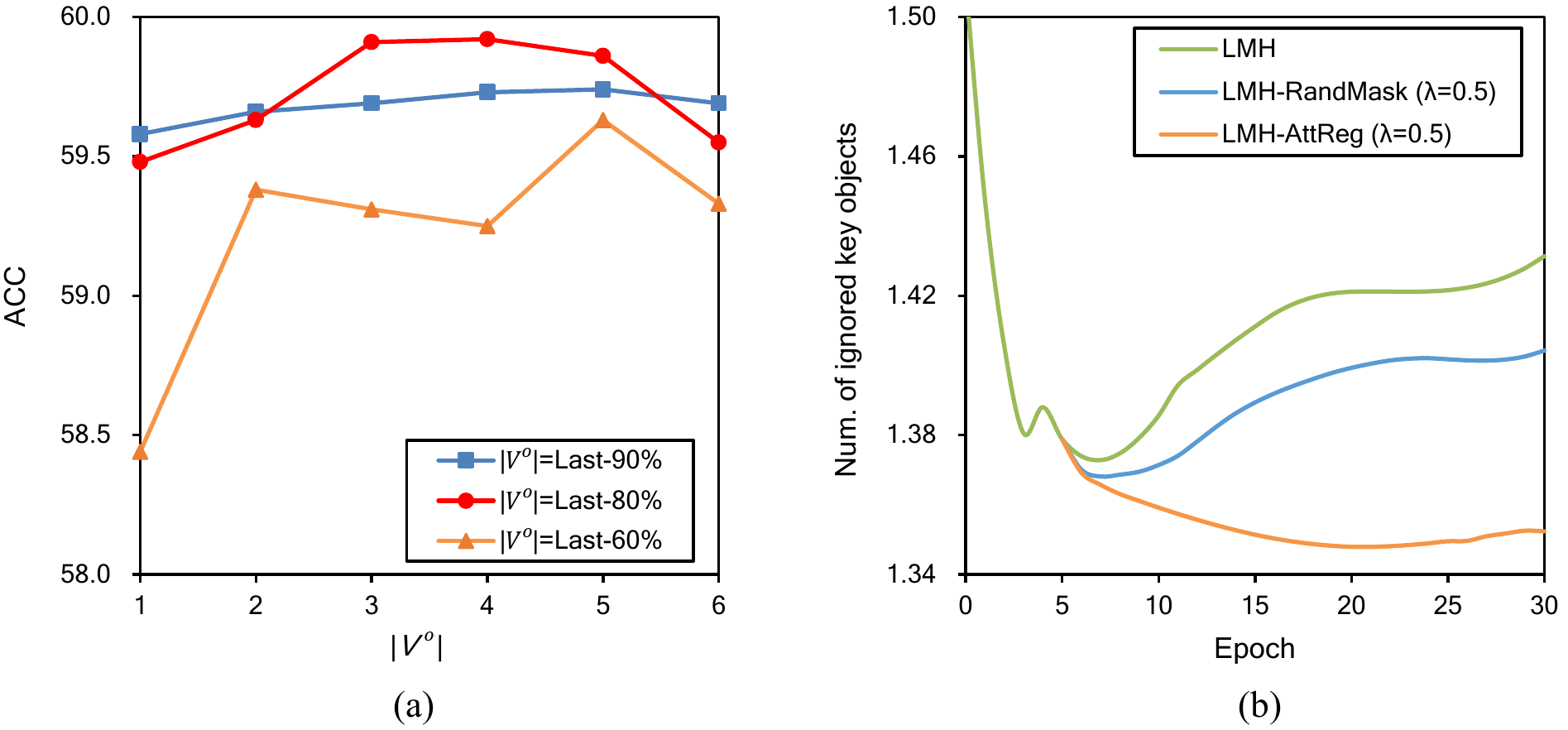} %
	\caption{(a) The ACC curves with respect to different sizes of $V^*$ and {the proportion of the deemed ignored objects $V^o$}. All results are obtained from the LMH-AttReg. (b) {The number of ignored key objects regarding the training epochs of LMH, LMH-RandMask, and our LMH-AttReg. Note that the RandMask and AttReg is introduced to fine-tune the backbone LMH after 5 epochs.}
	}
	\label{para_analysis}
\end{figure*}

\bfstart{The Size of Key Objects $V^*$} To evaluate the influence of $|V^*|$, we varied it from 1 to 6 to train our model with different settings on the VQA-CP v2 dataset and reported the results in Figure \ref{para_analysis} (a). 
It can be observed that the model accuracy obtains enhancement with a larger size of $V^*$ but deteriorates at some points, \textit{e.g.}, $|V^*|=6$.
The reason is that the image often contains certain number of key objects, and therefore the performance is promoted when more key objects are considered.
Nonetheless, the objects with lower similarities with QA pairs can also be included when $|V^*|$ is too large. This would introduce noise into the identified key objects and deteriorate AttReg's effect.

%Thus, we compared the performance variation of different proportion of $N$\% to evaluate the influence of $|V^o|$.
\bfstart{The Size of Deemed Ignored Objects $V^o$} 
As aforementioned, we took the 
{objects with attention weights in the last-$N$\% as the ignored objects $V^o$. Thus, we quantify different $N$ values to examine the effects of $|V^o|$, and the results are illustrated in Figure \ref{para_analysis} (a).
	It can be found that our AttReg performs better when $V^o$ includes more objects with low weights, \textit{e.g.}, last-90\%.
	This is because a majority of low-rank objects are less influential and should be deemed as ignored objects.
	As such, when $V^o$ is larger,
} the missed ignored objects in the locating process would be fewer and our AttReg can work better under this situation.

\begin{table*}[t]
	%	\caption{Global caption}
	\begin{minipage}{.47\linewidth}
		\caption{Performance comparison between the proposed method using different training strategies. \textit{w/o finetuning} means that the AttReg is introduced to regularize the backbone in an end-to-end way.
		}\smallskip
		\centering
		\resizebox{1.0\columnwidth}{!}{
			\smallskip\begin{tabular}{l c c c c}
				\toprule[1pt]
				
				Method & All & Yes/No & Number & Other\\
				
				Updn & 40.09 & 42.16 & 12.36 & 46.61\\
				\cmidrule{2-5}
				UpDn-AttReg (w/o finetuning) & 40.31 & 42.35 & \textbf{12.82} & \textbf{46.78}\\
				UpDn-AttReg (finetuning) & \textbf{46.85} & \textbf{69.34} & 12.44 & 44.51\\
				
				\midrule
				LMH & 52.99 & 72.02 & 39.24 & 46.79\\
				\cmidrule{2-5}
				LMH-AttReg (w/o finetuning) & 57.85 & 83.42 & 50.47 & 46.47\\
				LMH-AttReg (finetuning) & \textbf{59.92} & \textbf{87.28} & \textbf{52.39} & \textbf{47.65}\\
				\bottomrule[1pt]
			\end{tabular}
			
		}
		\label{finetune_ablation}
	\end{minipage}\hfill
	\begin{minipage}{.48\linewidth}
		\caption{The influence of the threshold $\sigma$ on the VQA-CP v2 test. \textit{Ratio} represents the proportion of training samples containing identified key objects.}\smallskip
		\centering
		\resizebox{1.0\columnwidth}{!}{
			\smallskip\begin{tabular}{l  c c  c c c c}
				\toprule[1pt]
				%			\hline
				& $\sigma$ & Ratio & All & Yes/No & Number & Other\\
				%			\hline \hline
				%			Updn &-&-& 40.09 & 42.16 & 12.36 & 46.61\\
				%			\cmidrule{2-5}
				\midrule
				\multirow{3}{*}{UpDn-AttReg} & 0.6 & 80\% & \textbf{46.85} & \textbf{69.34} & \textbf{12.44} & 44.51\\
				& 0.8 & 70\% &45.05 & 64.95 & 11.94 & 43.70\\
				& 1.0 &  20\% & 44.01 & 59.37 & 12.31 & \textbf{44.66}\\
				
				%			\hline \hline
				\midrule
				
				\multirow{3}{*}{LMH-AttReg} & 0.6 & 80\% & \textbf{59.92} & \textbf{87.28} & \textbf{52.39} & \textbf{47.65}\\
				& 0.8 & 70\% &58.41 & 83.65 & 50.64 & 47.33\\
				& 1.0 &  20\% & 54.41 & 77.66 & 37.26 & 46.88\\
				
				\bottomrule[1pt]
				%			\hline
			\end{tabular}
			
		}
		\label{preprocess_ablation}
	\end{minipage} 
\end{table*}

%
%\begin{table*}[t]
%	\caption{The influence of the threshold $\sigma$ on the VQA-CP v2 test. \textit{Ratio} represents the proportion of training samples containing identified key objects.}\smallskip
%	\centering
%	\resizebox{0.65\columnwidth}{!}{
%		\smallskip\begin{tabular}{l  c c  c c c c}
%			\toprule[1pt]
%			%			\hline
%			& $\sigma$ & Ratio & All & Yes/No & Number & Other\\
%			%			\hline \hline
%			%			Updn &-&-& 40.09 & 42.16 & 12.36 & 46.61\\
%			%			\cmidrule{2-5}
%			\midrule
%			\multirow{3}{*}{UpDn-AttReg} & 0.6 & 80\% & \textbf{46.85} & \textbf{69.34} & \textbf{12.44} & 44.51\\
%			& 0.8 & 70\% &45.05 & 64.95 & 11.94 & 43.70\\
%			& 1.0 &  20\% & 44.01 & 59.37 & 12.31 & \textbf{44.66}\\
%			
%			%			\hline \hline
%			\midrule
%			
%			\multirow{3}{*}{LMH-AttReg} & 0.6 & 80\% & \textbf{59.92} & \textbf{87.28} & \textbf{52.39} & \textbf{47.65}\\
%			& 0.8 & 70\% &58.41 & 83.65 & 50.64 & 47.33\\
%			& 1.0 &  20\% & 54.41 & 77.66 & 37.26 & 46.88\\
%			
%			\bottomrule[1pt]
%			%			\hline
%		\end{tabular}
%		
%	}
%	\label{preprocess_ablation}
%\end{table*}

\bfstart{The Training Strategy}
{
	For a further analysis of AttReg, we compared the effect of AttReg when using different training strategies, and the results are reported in Table \ref{finetune_ablation}. 
	It can be found that the AttReg performs best when using a finetuning way.
	The reason to this is that at the beginning of training, the visual attention module has not been trained favorably.
	%That is, whatever the input is the average of image features or the weighted one obtained via visual attention, they are nearly the ``same'' for the model, \textit{i.e.}, very close model performance.  
	As a result, it may introduce noises for the model learning if we apply our AttReg too early.
}

{
	\bfstart{The Threshold $\sigma$} Table \ref{preprocess_ablation} presents the performance of AttReg when using different threshold $\sigma$ to filter key objects. The results show that AttReg achieves better performance as $\sigma$ becomes lower. This is because when the filtering condition is relatively looser, the objects whose category are more roughly related to the QA pair can be remained, e.g., \textit{computer} (object category) and \textit{laptop} (noun in QA). 
	Meanwhile, in this situation, AttReg can utilize more samples to regularize the model.
}

\subsection{Effectiveness in Improving Visual Attention}
\bfstart{Quantitative Evaluation}
%Figure \ref{para_analysis} (b) shows how the number of ignored key objects changes with the increase of training steps. It can be seen that the number of ignored key objects assigned by the visual attention in the baseline LMH continuously {increases} with more training iterations after epoch {5}. In contrast, this number is decreasing when the visual attention is guided by our AttReg, especially when the regularization strength of AttReg becomes stronger. This demonstrates the effectiveness of AttReg in regularizing the visual attention to focus more on the ignored key objects.
Figure \ref{para_analysis} (b) shows how the number of ignored key objects changes {regarding} the increase of training steps. It can be seen that the number of ignored key objects assigned by the visual attention in the baseline LMH continuously {increases} with more training iterations after epoch {5}. In contrast, this number is decreasing when the visual attention is guided by our AttReg, {which demonstrates the effectiveness of AttReg} in regularizing the visual attention to focus more on the ignored key objects.
{In addition, we further introduced a new baseline (\ie, RandMask) for comparison, which can be regarded as a variant of our AttReg. That is, RandMask directly composes new samples by randomly masking images, while AttReg follows the human-based visual clues. 
As shown in Figure \ref{para_analysis} (b), we can observe that the number of ignored key objects continuously increases when RandMask is employed, yielding the same trend as the baseline LMH but the opposite trend as our AttReg. This result further highlights the superiority of our AttReg to the simple regularization, \ie, the capability of guiding model attention.}

To more intuitively understand the improvement of visual attention, we conducted ablation studies and calculated the performance gap for the backbone model with or without the visual attention module. As shown in Table \ref{att_ablation}, we can observe that the performance gap becomes larger if the visual attention is regularized by our AttReg, showing that the visual attention module plays a more pivotal role in model performance enhancement.

\begin{table*}[t]
	\caption{Influence of the visual attention module on VQA-CP v2 and VQA v2 datasets. Gap$\Delta$ denotes the performance variation in comparison with the baseline.}\smallskip
	\centering
	\resizebox{0.55\columnwidth}{!}{
		\smallskip\begin{tabular}{l c c c c c}
			\toprule[1pt]
			
			& \multirow{2}{*}{Method} & \multicolumn{2}{c}{VQA-CP v2} & \multicolumn{2}{c}{VQA v2}\\
			\cmidrule(lr){3-4} \cmidrule(l){5-6}
			& & All & Gap$\Delta$ & All & Gap$\Delta$ \\ \midrule
			
			\multirow{2}{*}{UpDn} & Baseline & 40.09 & - & 63.77 & - \\ 
			& w/o VAtt & 29.62 & 10.47 & 51.87 & 11.9 \\ 
			\midrule
			\multirow{2}{*}{UpDn-AttReg}  & Baseline & 46.85 & - & 64.13 & - \\
			& w/o VAtt & 35.34 & 11.51 & 51.88 & 12.25 \\

			\bottomrule[1pt]
		\end{tabular}
	}
	\label{att_ablation}
\end{table*}

\begin{figure*}
	[t]
	\centering
	\includegraphics[width=0.95\columnwidth]{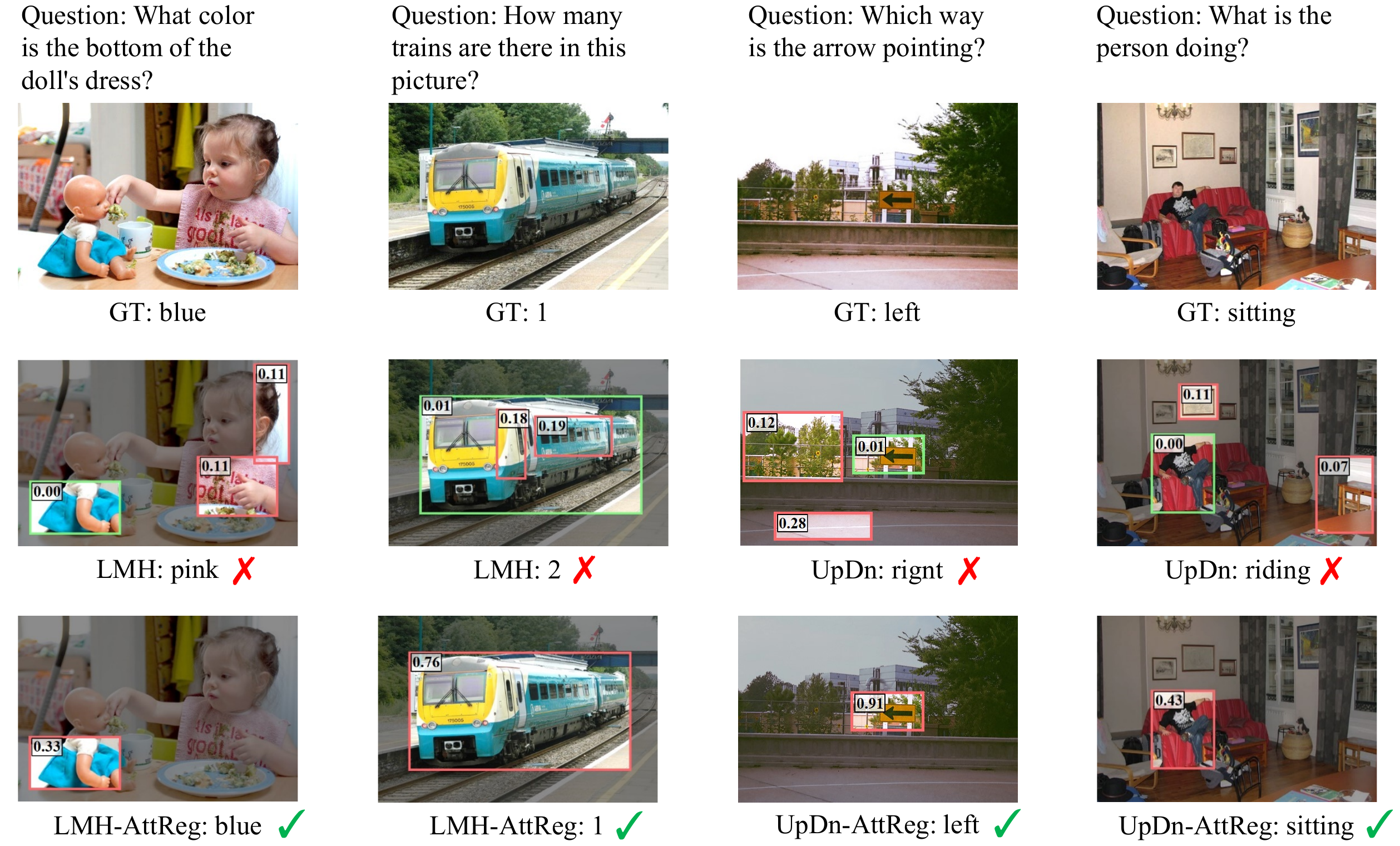} %
	\caption{Visualization of two backbone models with and without AttReg. 
		The green boxes denote the ignored key objects and the pink ones represent the objects with highest attention weights. The value around the bounding box is the visual attention weight to the given object. }
	\label{visualization}
\end{figure*}

\bfstart{Qualitative Evaluation}
To better illustrate the effectiveness of AttReg, we visualized the attention maps generated by the backbones with and without our AttReg, and exhibited the results in Figure \ref{visualization}. In all cases, the backbone model shifts more attention to the ignored key objects after introducing our AttReg. 
And the model is also promoted to predict the right answer with the correct visual grounding. Take the left one as an example, with our AttReg, the attention weight of the ignored key object \textit{dress} grows from 0.00 to 0.33, and becomes the most influential one among all objects in the image, which further help to yield the correct answer of \textit{blue}.

\subsection{Visual Attention Faithfulness Exploration}
\label{Sec_VAttExp}
One intuitive property in visual attention is that the image regions with larger attention weights should contribute more to model {predictions}, since these regions represent where the model focuses on when making decisions. Following \cite{Baseline_AttSupv(Argument):HINT}, we define this property as the faithfulness of visual attention.
To justify it, we conduct occlusion studies and quantify the contribution of {image regions} via two measurement: performance change and prediction variation. 
%To validate this assumption, we run occlusion studies \cite{Tech:GradCAM} and measure the contribution of regions with different attention weights from two aspects: performance change and prediction variation. 
%In particular, we further compare the faithfulness of visual attention with gradient-based explanations to testify the superiority of visual attention. The experiments are conducted on two datasets, VQA v2 and VQA-CP v2. 

\begin{figure*}
	[t]
	\centering
	\includegraphics[width=0.75\columnwidth]{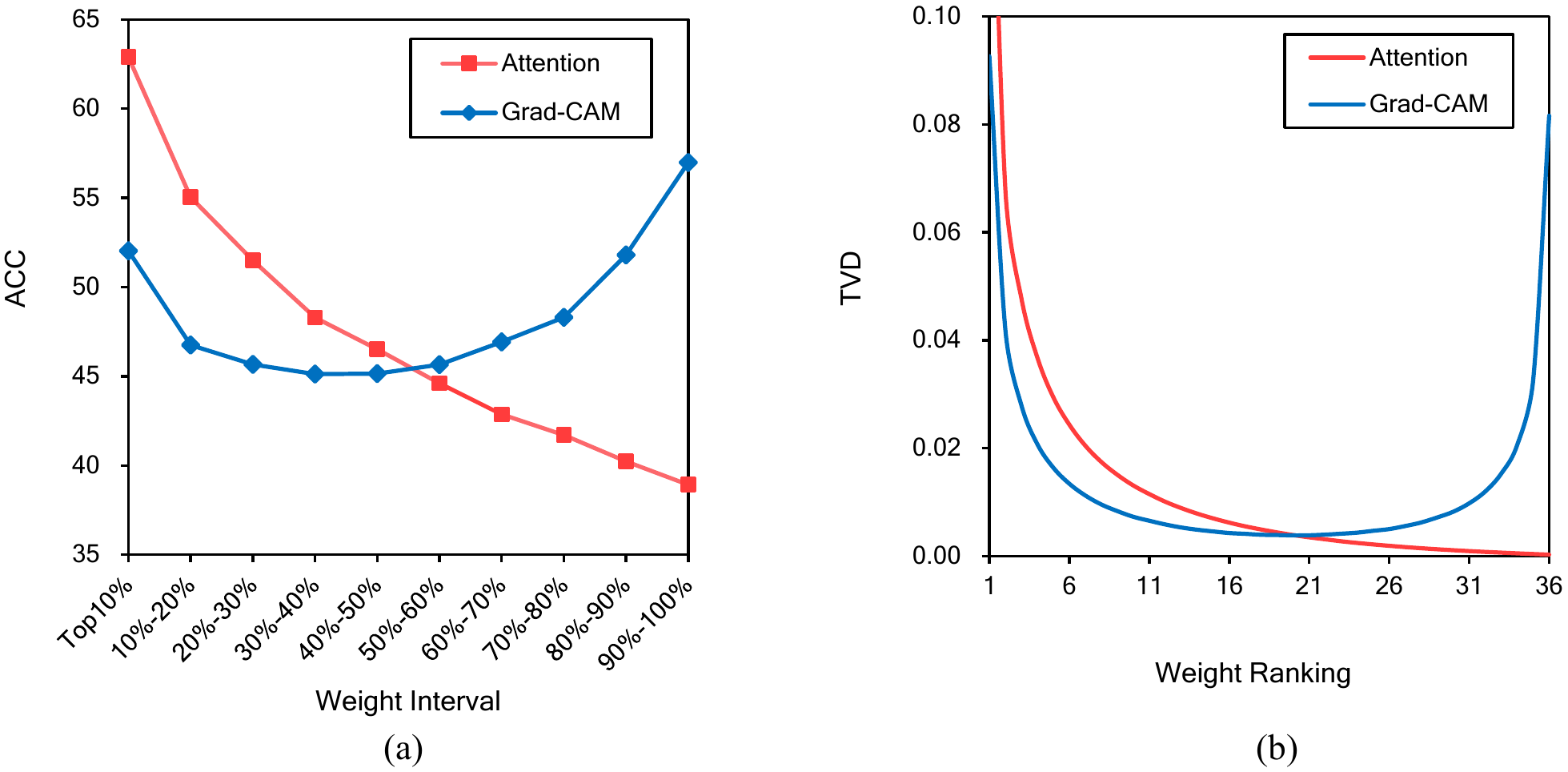} % Reduce the figure size so that it is slightly narrower than the column. Don't use precise values for figure width.This setup will avoid overfull boxes. 
	\caption{Faithfulness evaluation. (a) Performance curve regarding the weight intervals. (b) The TVD value with respect to weight ranking. 
	}
	\label{att_faith_fig}
	
\end{figure*}

%As shown in Table \ref{att_faith_fig} (a), the model performance continuously declines as the attention weights of the maintained regions decrease.
%Moreover, we further examine the model performance when only maintaining the 10\% or 20\% lowest attention weights of image regions. As can be expected, the performance of the model largely drops compared with the baseline. 
%Moreover, when only maintaining the 10\% or 20\% lowest attention weights of image regions, the performance of these two models largely drops compared with the baseline. 
%These preliminary observations are exciting since they are corresponding to our previous assumptions that image regions with larger attention weights are more responsible for model decisions.
%According to these preliminary observations, we can find that image regions with larger attention weights are more responsible for model decisions.
\bfstart{Contribution to Model Performance}
We empirically characterize visual attention weights as the contribution of image regions towards model performance, and measure the performance change when using different portions of image features,
{\textit{e.g.}, image features with 10\%-20\% highest attention scores\footnote{Note that the image features we used are composed of 36 object proposal features extracted by Faster R-CNN.}.
	In addition, we also leverage the same setting to testify the faithfulness of the Grad-CAM, and all the results are illustrated in Figure \ref{att_faith_fig} (a). 
	We can see that for the curve related to attention, the model performance continuously declines as the attention weights of the maintained regions decrease. 
	In contrast, by employing Grad-CAM as the visual grounding, the model performance drops first but then improves with the descending of the maintained weights. Further, the model performance of maintaining the lowest gradient weights even superior to the one with the highest gradient weights, which is confusing since regions with larger gradients in Grad-CAM should contribute more to the model performance. Instead, the visual attention weight shows a more acceptable consistency in model performance change. 
	
}

%This finding further verifies our previous observation in Table \ref{att_faith_table} that the visual attention is more faithful than Grad-CAM pertaining to visual grounding. 
%These findings further confirm our previous conclusions in Table \ref{att_faith_table}, revealing the better faithfulness of visual attention to model decisions.
%Instead, the model performance initially descend but begin to increase when the region's gradient weight deteriorates to one point. 

%Note that the Grad-CAM one is obtained by using answers with highest predicted probabilities.
% The Grad-CAM one is derived from the  answers with highest predicted probabilities.

\bfstart{Contribution to Model Prediction}
To further study the relationship between the attention weight and the contribution of the image region to model prediction, we remove each region sequentially according to its assigned weight value and observe the prediction variation.
Specifically, the region contribution is quantified through the prediction variation with and without the current image region, which is expressed by the Total Variation Distance (TVD) \cite{Argument:AttIsNotExp},
\begin{equation}
{\rm TVD} (p_1, p_2) = \frac{1}{2} \sum_{i} |p_{1i} - p_{2i}|,
\end{equation}
where $p_{1}$ and $p_{2}$ represent two different sets of predicted scores for each answer, respectively. A higher TVD denotes that the tested region is more influential for answer prediction.

We have computed the TVD for each image region with respect to its ranked weight and plotted the results in Figure \ref{att_faith_fig} (b). It can be found that the TVD is monotonously decreasing along the descending of the ranked attention weight, which demonstrates the tight correlation between the attention weight and region contribution towards answer prediction. 
In contrast, many regions with a very low ranking of Grad-CAM weights (\textit{i.e.}, ranking 26 to 36) yield a strong influence for answer prediction, which is confusing and violates the visual-grounding ability of Grad-CAM to some extent. 
{Based on the above experiments, we can conclude that the visual attention is more faithful than Grad-CAM pertaining to visual grounding in VQA.}
%Nevertheless, some key objects are still ignored by the visual attention as observed in our experiment, as shown in Figure \ref{para_analysis} (b). Inspired by this, we intend to develop a more advanced and pluggable approach to regularizing the visual attention, which can effectively promote the backbone model performance.

\section{Conclusion and Future Work}
\label{Sec_Conclusion}
In this work, we present a model agnostic visual attention regularization approach, \textit{i.e.}, AttReg, to guide the attention learning in VQA. AttReg has been applied to two strong baselines and significantly improves the backbone model performance over the VQA-CP v2 and VQA-CP v1 datasets. As a by-product, AttReg achieves a new state-of-the-art performance on VQA-CP v2.
In addition, we also empirically study the faithfulness of visual attention in VQA. The experimental results have demonstrated that { the visual attention obviously outperform the Grad-CAM in terms of visual grounding. }

In the future, we will extend our approach to other tasks which are also hindered by the unsupervised attention learning problems, \textit{e.g.}, image captioning.

%\section{Acknowledgments}
%\begin{acks}
%\end{acks}

\bibliographystyle{ACM-Reference-Format}
\bibliography{bibliography}

%\appendix

\end{document}